%% file: main.tex
\documentclass[twocolumn,pagebackref,breaklinks,colorlinks,allcolors=iccvblue]{fairmeta}

\input{preamble}

\usepackage{url}
\usepackage{makecell}

\usepackage{epsfig}
\usepackage{amsmath}
\usepackage{amssymb}
\usepackage{fontawesome5}

\usepackage{multirow, booktabs, graphicx, xcolor, colortbl, arydshln, tikz}
\usetikzlibrary{tikzmark}
\usepackage{siunitx} 
\usepackage{mdframed}
\usepackage{tablefootnote} 

\newcolumntype{g}{>{\columncolor{gray!30}}c}

\usepackage{tcolorbox}
\tcbuselibrary{most}

\definecolor{citecolor}{HTML}{0071bc}
\hypersetup{
    colorlinks=true,
    citecolor=citecolor,
    filecolor=citecolor,
    linkcolor=citecolor,
    urlcolor=citecolor
}

\tcbuselibrary{skins}

\definecolor{userbg}{RGB}{245, 245, 245}
\definecolor{userborder}{RGB}{210, 229, 255}
\definecolor{userfont}{RGB}{0, 0, 0}

\tcbset{
  enhanced,
  boxrule=1pt,
  colback=userbg,
  colframe=userborder,
  fonttitle=\bfseries,
  coltitle=userfont,
  boxsep=5pt,
  arc=5pt,
  outer arc=5pt,
  left=10pt,
  right=10pt,
  top=2pt,
  bottom=2pt,
  attach boxed title to top left={yshift=-0.25\baselineskip-1mm,xshift=2mm},
  boxed title style={
    colback=userborder,
    sharp corners,
    rounded corners=northeast,
    arc=3pt,
    outer arc=3pt,
    boxrule=0pt,
    boxsep=2pt,
  },
}
\usepackage{pgf}
\usepackage{adjustbox}
\usepackage{enumitem}
\usepackage{array}
\definecolor{listcolor}{RGB}{50,120,230}

\definecolor{w_1}{RGB}{66,138,244}
\definecolor{w_2}{RGB}{73,144,245}
\definecolor{w_3}{RGB}{79,148,246}

\usepackage{caption}

\newcounter{researchquestion}

\newcommand{\researchquestion}[2][]{
  \vspace{0.8em}
  \refstepcounter{researchquestion}
  \begin{tcolorbox}[
    enhanced,
    colback=blue!5,
    colframe=blue!70!black,
    fonttitle={\fontsize{10.5pt}{12.8pt}\selectfont\bfseries\color{blue!20!black}},
    title=Question \theresearchquestion,
    toprule=1.5pt,
    bottomrule=0.8pt,
    leftrule=0.8pt,
    rightrule=0.8pt,
    left=6pt,
    right=6pt,
    top=6pt,
    bottom=6pt,
    boxsep=3pt
  ]
  \normalsize #2
  \end{tcolorbox}
  \ifx\\#1\\\else\label{rq:#1}\fi
  \vspace{0.5em}
}

\title{GoViG: Goal-Conditioned Visual Navigation Instruction Generation via Multimodal Reasoning}

\author[1,*]{Fengyi~Wu}
\author[1,*]{Yifei~Dong}
\author[1]{Yilong~Dai}
\author[1]{Guangyu~Chen}
\author[1]{Qifeng~Wu}
\author[1]{Huiting~Huang}
\author[2]{Hang~Wang}
\author[3]{Qi~Dai}
\author[4]{Alexander~G.~Hauptmann}
\author[1,\dagger]{Zhi-Qi~Cheng}

\affiliation[1]{University~of~Washington}
\affiliation[2]{The~Hong~Kong~Polytechnic~University}
\affiliation[3]{Microsoft~Research}
\affiliation[4]{Carnegie~Mellon~University}

\contribution[*]{Equal Contributions}
\contribution[\dagger]{Corresponding author.}

\abstract{
We introduce \textit{Goal-Conditioned Visual Navigation Instruction Generation} (GoViG), a new task that aims to generate contextually coherent navigation instructions solely from egocentric visual observations of initial and goal states. Unlike prior work relying on structured inputs, such as semantic annotations or environmental maps, GoViG exclusively leverages raw egocentric visual data, improving adaptability to unseen and unstructured environments. Our method
addresses this task by decomposing it into two interconnected subtasks: (1) navigation visualization, predicting intermediate visual states bridging the initial and goal views; and (2) instruction generation, synthesizing coherent instructions grounded in observed and anticipated visuals. Both subtasks are integrated within an autoregressive multimodal LLM trained with tailored objectives to ensure spatial accuracy and linguistic clarity. Furthermore, we introduce two multimodal reasoning strategies, one-pass and interleaved reasoning, to mimic incremental human navigation cognition. To comprehensively evaluate our method, we propose the \textit{R2R-Goal} dataset, combining diverse synthetic and real-world trajectories. Empirical results demonstrate significant performance improvements over state-of-the-art methods in BLEU-4 and CIDEr scores along with robust cross-domain generalization.
}

\date{August 13, 2025}

\project{\href{https://github.com/F1y1113/GoViG}{\sffamily \fontsize{8.8pt}{11pt}\selectfont \texttt{https://github.com/F1y1113/GoViG}}}

\begin{document}
\maketitle

\section{Introduction}

Generating natural language navigation instructions from egocentric visual observations remains a critical yet underexplored area within embodied AI. While Vision-and-Language Navigation (VLN) research has largely focused on language grounding,~training agents to interpret and execute human instructions~\cite{anderson2018vision, fried2018speakerfollowermodelsvisionandlanguagenavigation},~the inverse challenge of \textit{instruction generation} is relatively understudied. Effective instruction generation is crucial for practical applications such as aiding visually impaired users, facilitating seamless human-agent collaboration, and guiding navigation in hazardous or unfamiliar environments~\cite{zhang2024visionsurvey}.

\begin{figure}[!t]
\vspace{0.1in}
\setlength{\abovecaptionskip}{2pt}
\setlength{\belowcaptionskip}{0pt}
\centering
\includegraphics[width=\linewidth]{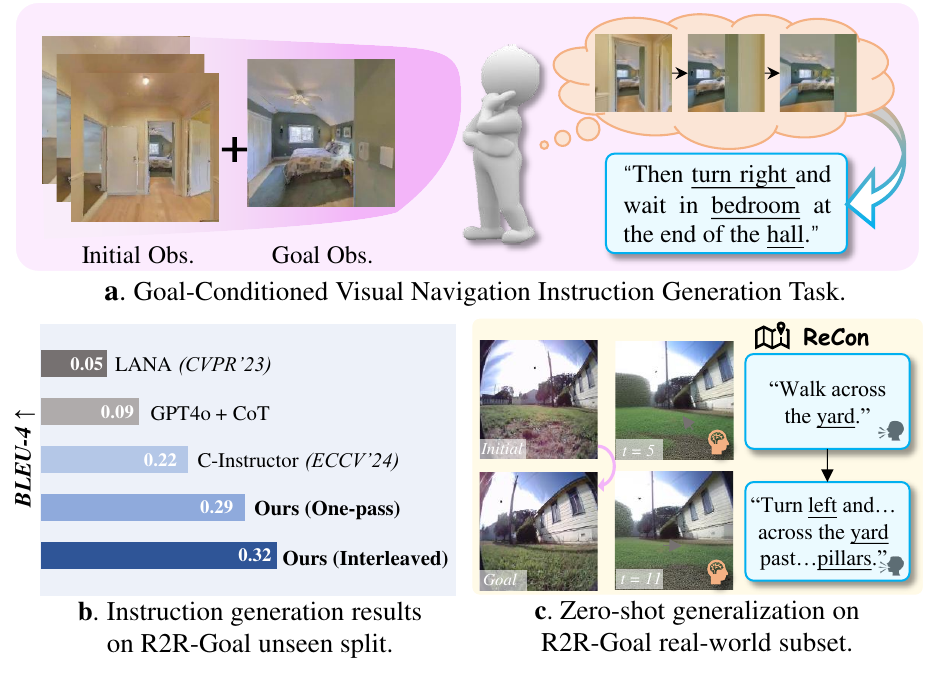}
\caption{\small (a) Goal-Conditioned Visual Navigation Instruction Generation (GoViG): generating instructions from egocentric initial and goal views. (b) Results on the R2R-Goal. (c) Zero-shot generalization to real-world scenarios.}
\label{fig:task_overview}
\vspace{-0.2in}
\end{figure}

Current methods for instruction generation predominantly depend on \textit{privileged inputs}, including semantic maps, landmark annotations, and panoramic views, limiting their applicability beyond controlled and structured scenarios~\cite{fan2024navigation, wang2023lanalanguagecapablenavigatorinstruction, Kong_2024}. Alternatively, some approaches simplify visual data into textual summaries, inadvertently discarding essential spatial and semantic information intrinsic to raw visual observations~\cite{fan2024navigation, wang2022less, zeng2023kefaknowledgeenhancedfinegrained}. Such oversimplifications impede an agent's capability for accurate reasoning and generalization in novel contexts.

Recent advancements in multimodal large language models (MLLMs), such as LLaVA~\cite{liu2023llava}, GPT-4o~\cite{hurst2024gpt4o}, and Gemini~\cite{gemini2,comanici2025gemini}, have demonstrated remarkable proficiency in vision-language tasks. Nevertheless, these models generally lack explicit mechanisms for coherent visualization and seldom incorporate iterative mental simulation strategies utilized by humans during route planning~\cite{chen2023see, zhang2023multimodal}. Consequently, instructions generated by existing MLLMs frequently suffer from a lack of contextual precision and temporal consistency.

To address these limitations, we introduce \textit{Goal-Conditioned Visual Navigation Instruction Generation} (GoViG), a novel task aiming to generate precise and contextually coherent navigation instructions using only egocentric visual observations from initial and goal viewpoints (Fig.\ref{fig:task_overview}(a)). Unlike previous approaches, GoViG entirely eliminates reliance on privileged inputs, significantly enhancing the method's generalization capability across diverse and unseen environments (Fig.\ref{fig:task_overview}(b)-(c)).

\begin{figure*}[!t]
\setlength{\abovecaptionskip}{-1pt}
\setlength{\belowcaptionskip}{0pt}
\centering
\includegraphics[width=\linewidth]{ 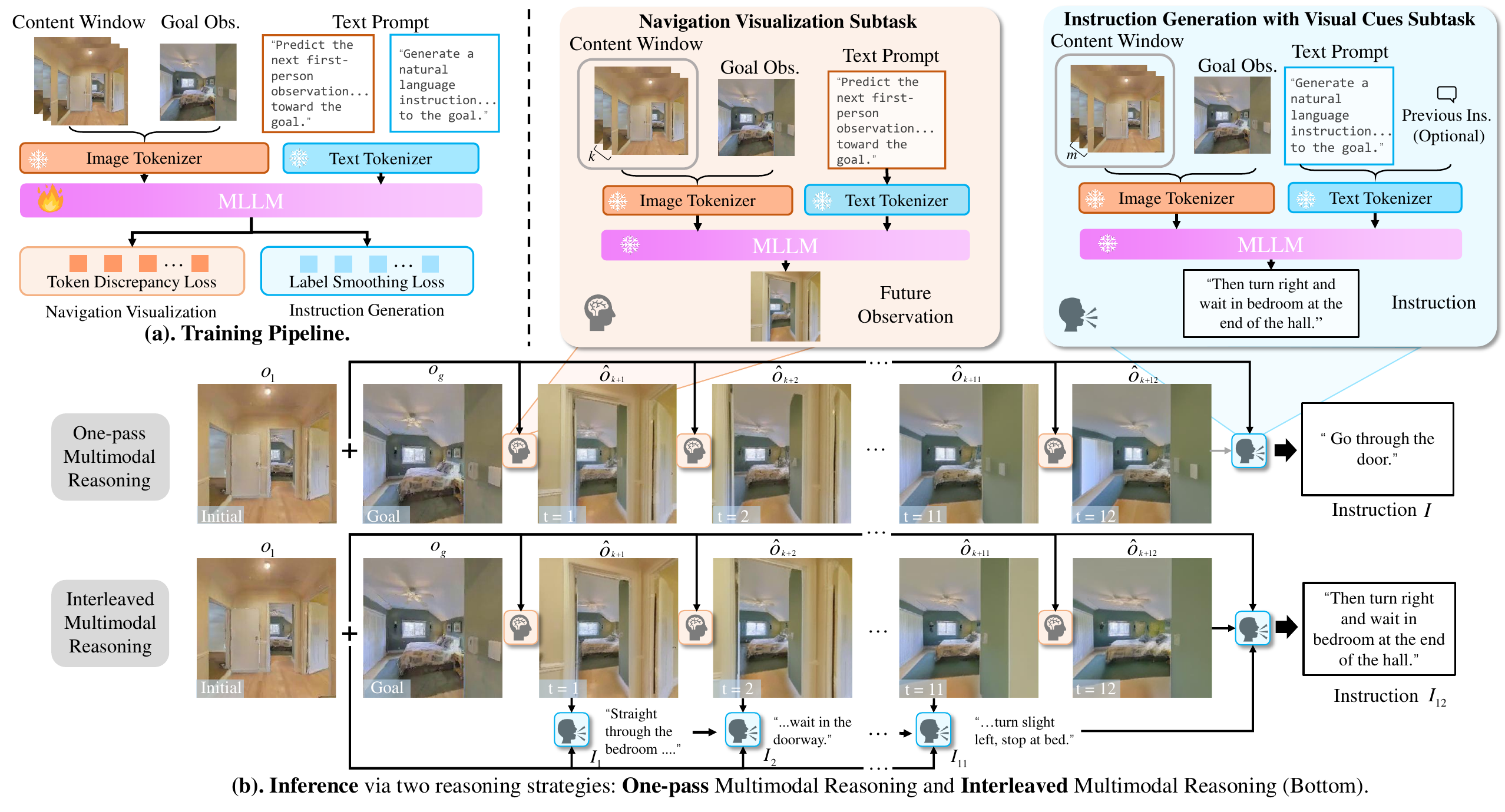}
\vspace{-0.1in}
\caption{\small \textbf{Overview of our approach} to Goal-Conditioned Visual Navigation Instruction Generation (GoViG): (a) An autoregressive MLLM integrates Navigation Visualization and Instruction Generation subtasks via tailored training objectives. (b) Two inference-time multimodal reasoning strategies: One-pass and Interleaved, which enable coherent visual forecasting and instruction synthesis. The context size is set to one for clarity. [Zoom in for details.]}
\label{fig:method overview}
\vspace{-0.2cm}
\end{figure*}

Our approach systematically decomposes GoViG into two complementary subtasks: (1) \textit{Navigation Visualization}, predicting intermediate visual states to bridge initial and goal observations; and (2) \textit{Instruction Generation with Visual Cues}, synthesizing instructions grounded in observed and forecasted visual cues (Fig.~\ref{fig:method overview}(a)).~Both are integrated within an autoregressive MLLM, guided by carefully designed training objectives: a \textit{Token Discrepancy Loss},~which promotes accurate visual predictions, and a \textit{Label Smoothing Loss}, enhancing semantic robustness and linguistic fluency. This methodological synergy aligns closely with human spatial cognition, fostering robust and adaptable instruction generation.

Furthermore, we propose two multimodal reasoning strategies during inference: \textit{One-Pass Multimodal Reasoning}, leveraging global visual context for structured scenarios; and \textit{Interleaved Multimodal Reasoning}, iteratively refining visual predictions and linguistic instructions to emulate human adaptive navigation under uncertainty (Fig.~\ref{fig:method overview}(b)). These strategies enhance spatial accuracy, linguistic coherence and cross-domain generalization.

We extensively evaluate GoViG on proposed \textit{R2R-Goal} dataset, integrating synthetic trajectories from R2R-CE~\cite{krantz_vlnce_2020} and HA-R2R~\cite{dong2025ha} with real-world egocentric videos from GO Stanford~\cite{hirose2018gonet}, ReCon~\cite{shah2021rapid}, and HuRoN~\cite{hirose2023sacson}, each meticulously annotated with natural language instructions. Our interleaved reasoning strategy achieves superior performance with BLEU-4 (0.32) and CIDEr (0.20) scores on validation (Table~\ref{tab:sota_instruction_comparison}). Moreover, it attains a BLEU-4 of 0.27 in zero-shot cross-domain evaluations (Table~\ref{tab:sota_cross-domain_instruction_comparison}), showing robust generalization capabilities.

Our contributions can be summarized as follows:
\begin{enumerate}[leftmargin=0.8em,itemsep=1pt,topsep=1pt]
\item We formally propose Goal-Conditioned Visual Navigation Instruction Generation (GoViG), a new task generating precise navigation instructions solely from egocentric initial and goal observations, without privileged inputs (Sec.~\ref{sec:task}).
\item We systematically decompose GoViG into two subtasks: Navigation Visualization and Instruction Generation with Visual Cues, and integrate them within a unified autoregressive MLLM optimized with tailored training objectives (Sec.~\ref{sec.mllm_training}).
\item We introduce and evaluate two multimodal reasoning strategies (One-Pass and Interleaved) designed to enhance spatial accuracy and linguistic coherence through global and iterative visual-linguistic reasoning (Sec.~\ref{sec.mllm_inference}).
\item We release the R2R-Goal dataset, a comprehensive benchmark combining synthetic and real-world navigation scenarios. Extensive empirical evaluations validate our method's superior instruction generation performance and robust cross-domain generalization (Secs.~\ref{sec:dataset}, \ref{sec:experiment}).
\end{enumerate}

\section{Related Work}
\subsection{Navigation Instruction Generation}
Navigation instruction generation originates from cognitive science research examining human spatial cognition and culturally influenced route descriptions~\cite{lynch1964image, allen1997knowledge, vanetti1988communicating, hund2006getting}. Recent advancements in VLN have renewed interest in instruction generation, primarily for data augmentation~\cite{anderson2018vision, zhang2024visionsurvey}. Early computational approaches, such as Speaker-Follower~\cite{fried2018speakerfollowermodelsvisionandlanguagenavigation}, employed recurrent neural networks to generate instructions. Subsequent methods~\cite{tan-etal-2019-learning, 9879110, wang2025bootstrappinglanguageguidednavigationlearning} enhanced instruction quality but continued relying on structured inputs, such as semantic annotations, panoramic images, and environmental maps~\cite{fan2024navigation, Fan_2025_CVPR, gopinathan2024spatially, zeng2023kefaknowledgeenhancedfinegrained, cui2025generatingvisionlanguagenavigationinstructions, yan2024instrugenautomaticinstructiongeneration, zhao2025lafgrpoinsitunavigationinstruction, wang2025navraggeneratinguserdemand}, limiting their generalization to novel scenarios.

Moreover, contemporary approaches often preprocess visual inputs into intermediate representations, such as landmarks or commonsense knowledge, unintentionally discarding critical spatial and semantic details inherent to raw visual observations~\cite{Kong_2024, cui2025generatingvisionlanguagenavigationinstructions, gopinathan2024spatially, zeng2023kefaknowledgeenhancedfinegrained, wang2025navraggeneratinguserdemand}. In contrast, our method explicitly leverages raw egocentric visual observations through multimodal reasoning strategies, One-Pass and Interleaved, to directly embed visual cognition into the instruction generation process. Further comparisons are provided in Appendix.

\subsection{Multimodal Reasoning}
Recent multimodal large language models (MLLMs)~\cite{yang2023dawnlmmspreliminaryexplorations, anthropic2024claude3, gemini2,comanici2025gemini} have advanced visual-textual understanding significantly. Models like LLaVA~\cite{liu2023llava}, BLIP-2~\cite{li2023blip}, and VideoChat~\cite{li2023videochat} excel at multimodal comprehension, while generative frameworks~\cite{hong2022cogvideo, henschel2024streamingt2v}, have enhanced video synthesis. Integrated architectures such as GPT-4o~\cite{hurst2024gpt4o} further showcase sophisticated multimodal reasoning. Concurrently, Chain-of-Thought (CoT) reasoning~\cite{wei2022chain} has emerged as essential within multimodal reasoning frameworks. Foundational methods~\cite{chen2023see, zhang2023multimodal} structured reasoning explicitly around visual data. Subsequent research~\cite{wei2024mc, li2023intentqa, zhao2023antgpt}) extended CoT reasoning to zero-shot video understanding and egocentric activities. Recent studies~\cite{shao2024visual, zhou2024image, wu2024mind} advocate spatially coherent visual-textual inference. Inspired by these advances, our method integrates visualization and CoT-based linguistic reasoning, enabling coherent navigation instruction generation directly from egocentric visuals.

\subsection{World Models for Visual Generation}

World models~\cite{ha2018world} have become a central paradigm for learning predictive representations of environment dynamics~\cite{ding2024understanding}, evolving from compact recurrent structures to large-scale generative and multimodal systems. Early works~\cite{ha2018world, hafner2019dream,hafner2022masteringataridiscreteworld,hafner2024masteringdiversedomainsworld} employed RNN-based latent dynamics to capture temporal transitions. Transformer-based designs~\cite{assran2023self, bardes2024revisiting, karypidis2024dino, baldassarre2025back} introduced scalable attention mechanisms for richer spatio-temporal abstraction. Parallel efforts exploit LLMs to simulate dynamics~\cite{zhao2025drivedreamer,xing2025critiquesworldmodels, dong2025unified, dong2026language}, but they face modality misalignment, temporal inconsistency, and grounding challenges~\cite{ding2024understanding, dong2025large}. Inspired by this paradigm, our Navigation Visualization subtask adopts a lightweight world-model perspective within an autoregressive MLLM: it iteratively predicts intermediate egocentric observations to bridge the initial and goal states, enabling downstream instruction generation that is grounded in anticipated visual futures.

\section{Methodology}
\label{sec:method}
\subsection{Task Overview~\&~Multimodal~Reasoning}
\label{sec:task}
\noindent\textbf{Task~Formulation.}~We define Goal-Conditioned Visual Navigation Instruction Generation (GoViG) as the task of generating coherent natural language instructions to guide an agent towards a specified goal using solely egocentric visual observations. Specifically, given an initial visual sequence $\mathcal{O}=\{o_1,o_2,\dots,o_n\}$ and a goal observation $o_g$, where each $o_i,o_g\in\mathbb{R}^{H\times W\times3}$ denotes an RGB egocentric image, the objective is to produce an accurate navigation instruction $I$ that clearly delineates necessary steps for reaching the goal. Inspired by world model paradigm of predicting future states to support downstream decision-making, we systematically approach this task by decomposing it into two interconnected subtasks:

\begin{enumerate}[label=\textbullet,leftmargin=0.8em,itemsep=1pt,topsep=1pt]
    \item \textbf{Navigation Visualization.} Given a partial visual observation sequence $\mathcal{O}_V=\{o_1,o_2,\dots,o_k\}$ and the goal observation $o_g$, the model predicts the next visual observation $o_{k+1}$, incrementally bridging the gap between the initial and goal states through visual imagination.
    \item \textbf{Instruction Generation with Visual Cues.} Given a visual sequence $\mathcal{O}_I=\{o_1,o_2,\dots,o_m\}$ (where typically $m=k+1$), the goal observation $o_g$, and optionally an intermediate instruction $I_{\text{prev}}$, the model generates a coherent, contextually grounded instruction $I$ that articulates the sequential navigation steps towards the goal.
\end{enumerate}

\noindent To facilitate training, we implement an autoregressive MLLM as illustrated in Fig.~\ref{fig:method overview}(a), with tailored loss functions specifically designed for each subtask. We present the detailed model architecture and training procedures in Sec.~\ref{sec.mllm_training}.

\noindent \textbf{Multimodal Reasoning.} Unlike conventional approaches that directly translate visual inputs into textual instructions, our framework explicitly integrates structured visual reasoning to improve robustness and generalization. Specifically, we introduce two distinct multimodal reasoning strategies, illustrated in Fig.~\ref{fig:method overview}(b) and detailed in Sec.~\ref{sec.mllm_inference}:

\begin{enumerate}[label=\textbullet,leftmargin=0.8em,itemsep=1pt,topsep=1pt]
    \item \textbf{One-Pass Multimodal Reasoning.} Given an initial visual sequence $\mathcal{O}_{\text{init}} = \{o_1,\dots,o_k\}$ and a goal observation $o_g$, the model forecasts a complete trajectory $\hat{\mathcal{O}} = \{\hat{o}_{k+1},\dots,\hat{o}_{k+t}\}$ toward the goal. Subsequently, the navigation instruction $I$ is generated from selected representative frames in $\hat{\mathcal{O}}$, emphasizing holistic spatial context and global scene awareness.

    \item \textbf{Interleaved Multimodal Reasoning.} Starting from initial observations $\mathcal{O}_{\text{init}}$, the model iteratively alternates between forecasting next visual observation $\hat{o}_{k+t}$ and incrementally updating corresponding instruction $I_t$. This approach closely mimics incremental human cognitive processes, ensuring precise alignment between visual perception and linguistic instruction generation.
\end{enumerate}

\noindent Unlike conventional techniques relying on explicit coordinates, action labels, or semantic maps, our approach solely employs egocentric visual observations, enabling enhanced generalization to diverse, unknown environments and laying the groundwork for cross-domain applications.

\subsection{Construction of the R2R-Goal Dataset}

\label{sec:dataset}
To support GoViG task, we introduce \textbf{R2R-Goal dataset}.~This dataset integrates language instructions from existing R2R-CE~\cite{krantz_vlnce_2020} and HA-R2R~\cite{dong2025ha, li2024human} datasets and incorporates first-person observations from GO Stanford~\cite{hirose2018gonet}, ReCon~\cite{shah2021rapid} and HuRoN~\cite{hirose2023sacson} datasets as a dedicated real-world test subset.

\begin{figure}[!t]
\setlength{\abovecaptionskip}{1pt}
\setlength{\belowcaptionskip}{1pt}
    \centering
    \includegraphics[width=1\linewidth]{ 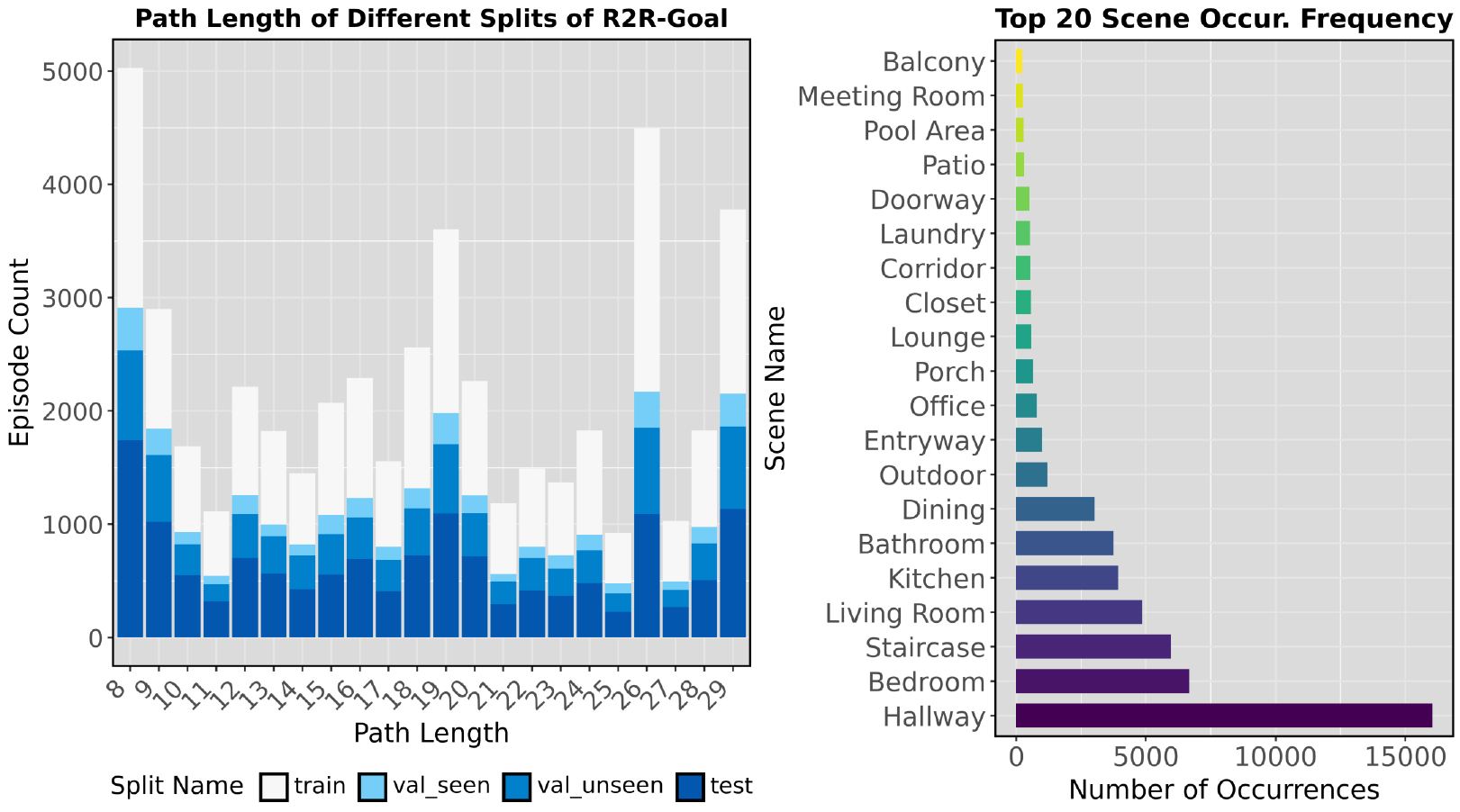}
    \caption{\small \textbf{R2R-Goal dataset statistics:} (\textit{left}) Distribution of trajectory lengths (8--29 steps) across training, validation (seen/unseen) and test splits; (\textit{right}) Top 20 scene categories ranked by frequency, showcasing coverage of diverse indoor and outdoor environments.}
    \label{fig:dataset}
\vspace{-0.3cm}
\end{figure}

To leverage R2R-CE and HA-R2R, we generate egocentric observation sequences and corresponding navigation paths using an A\textsuperscript{*}-based heuristic search in the HA-VLN simulator~\cite{dong2025ha}, using Qwen-VL-2.5~\cite{bai2025qwen2} to segment both visual observation and corresponding instructions into semantically coherent sub-scenes. All trajectory-instruction pairs are manually reviewed to verify spatial coherence and semantic alignment. Misaligned or ambiguous pairs are corrected or removed. This human-in-the-loop validation ensures diverse, high-quality navigation patterns.~As a result, this part of the R2R-Goal dataset consists of 74,737 trajectories, partitioned into training (48,490), validation~seen (3,573), validation~unseen (8,361), and testing (14,313) splits, with detailed statistics presented in Fig.~\ref{fig:dataset}.

For the real-world subset, we apply the same segmentation strategy to observation sequences obtained from the GO Stanford, ReCon, and HuRoN datasets. We then manually annotate a total of 1080 trajectories with corresponding natural language navigation instructions. Each trajectory in R2R-Goal retains an initial sequence of six egocentric observations and a final goal observation. These visual sequences, combined with the corresponding navigation instructions, constitute the inputs for our proposed task. Additional dataset construction and annotation details are included in the Appendix.

\subsection{Autoregressive MLLM Training}
\label{sec.mllm_training}
To integrate visual and linguistic reasoning within a unified framework, we employ an autoregressive multimodal Transformer based on the Chameleon architecture~\cite{team2024chameleon}. This design simultaneously addresses two complementary subtasks: Navigation Visualization and Instruction Generation with Visual Cues as illustrated in Fig.~\ref{fig:method overview}(a) and detailed in Fig.~\ref{fig:detail_training}, facilitating shared representation learning and robust multimodal interaction.

\noindent\textbf{Data Preparation \& Prompt Design.}~We construct training samples by pairing egocentric RGB image sequences with natural language navigation instructions. Each trajectory generates two types of training instances: (1) \textbf{Navigation Visualization}: each instance consists of a sequence of $k$ preceding visual frames alongside the goal frame, with the task being to predict the next frame. Visual observations are denoted by \texttt{<image>} tokens in multimodal prompts, with multiple samples extracted via a sliding temporal window. (2) \textbf{Instruction Generation with Visual Cues}: inputs consist of the initial frame, goal frame, and up to $m-1$ intermediate frames, embedded within an image prompt. The associated ground-truth instruction serves as the prediction target. Samples from both subtasks are interleaved in batches for joint optimization using a unified Transformer model. Prompt examples are provided in the Appendix.

\noindent\textbf{Multimodal Tokenization.}
To effectively integrate visual and textual modalities, our model employs two specialized tokenizers. The first is a vector-quantized (VQ) image tokenizer derived from~\cite{team2024chameleon}, discretizing images into sequences of visual tokens via a learned embedding codebook. The second is an optimized byte-pair encoding (BPE) tokenizer, following~\cite{team2024chameleon, tan-etal-2019-learning}, converting textual navigation instructions into discrete token sequences. Visual and textual token sequences are concatenated and jointly processed by a causal Transformer, promoting coherent multimodal representations.

\noindent\textbf{Subtask-Specific Training Objectives.}
To optimize our model for distinct characteristics of each subtask, we introduce tailored training objectives. At each iteration, our autoregressive MLLM jointly processes samples from either the navigation visualization or instruction generation subtask, producing logits across the unified vocabulary. The subtask-specific loss functions are detailed as follows:

To construct navigation visualization loss, we utilize Token Discrepancy Loss~\cite{li2025imagine} to encourage accurate visual forecasting. Given the ground-truth visual embedding $\text{emb}_i$ for token $i$ (out of total $n$ tokens in current image) and visual codebook embeddings $\mathcal{C}=\{\text{emb}_1, \dots, \text{emb}_N\}$ where $N$ denotes the total number of visual token vocabulary, the loss is computed as:
\begin{equation}
\setlength\abovedisplayskip{1pt}
\setlength\belowdisplayskip{1pt}
\mathcal{L}_{\text{vis}} = \sum\nolimits_{i=1}^{n} \mathrm{MSE}(\text{emb}_i, \mathcal{C}) \cdot P(t_i),
\end{equation}
where $\!\mathrm{MSE}(emb_i, \mathcal{C})\!\!\in\!\!\mathbb{R}^{1 \!\times\!N}\!\!$~is similarity vector containing distances between $emb_i $ and all codebook entries. $\!P(t_i)\!\in\!\mathbb{R}^{1\!\times\!N}\!\!$~denotes predicted probability distribution for visual tokens at position $i$.

For instruction generation loss design, we apply a label smoothing cross-entropy loss. Let $y_i$ denote the ground-truth text token at position $i$ in the target instruction sequence, the loss is defined as:
\begin{equation}
\setlength\abovedisplayskip{1pt}
\setlength\belowdisplayskip{1pt}
\mathcal{L}_{\text{ins}} = -\sum\nolimits_{i}\sum\nolimits_{v\in\mathcal{V}} q_v(y_i)\log P_v(y_i),
\end{equation}
where $P_v(y_i)$ is predicted probability for token $v$ at position $i$, and $q_v(y_i)$ represents smoothed distribution around $y_i$ within text vocabulary $\mathcal{V}$, applying smoothing factor $\epsilon$ to non-ground-truth tokens.

To stabilize training, we implement an input-label concatenation strategy, masking inputs (with $-100$ labels) so loss computation focuses exclusively on prediction targets. During training, tokenizers remain frozen, and only Transformer parameters are updated via an autoregressive next-token prediction objective.

\begin{figure*}[!t]
\setlength{\abovecaptionskip}{1pt}
\setlength{\belowcaptionskip}{0pt}
    \centering
    \includegraphics[width=0.97\linewidth]{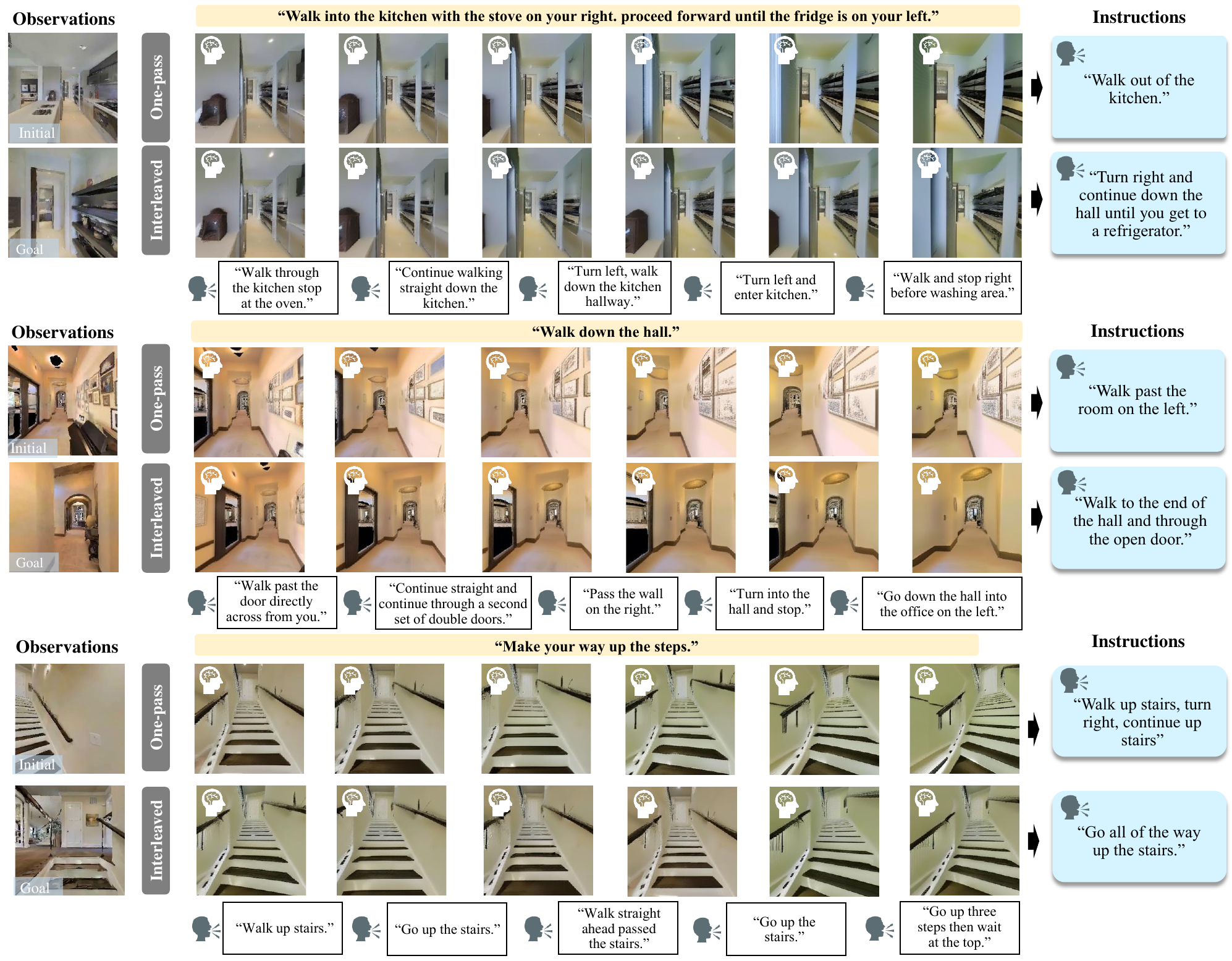}
    \caption{\small \textbf{Qualitative examples} of Navigation Visualization and Instruction Generation on R2R-Goal validation unseen split.~Both our Multi-modal reasoning strategies perform well on diverse unseen scenes such as stairs and kitchens.}
    \label{fig:unseen_result}
    \vspace{-0.2cm}
\end{figure*}

\subsection{Multimodal Reasoning Strategies}
\label{sec.mllm_inference}
During inference, we leverage the trained MLLM $F_\Theta$, with fixed parameters, employing two structured multimodal reasoning strategies: One-Pass and Interleaved. Both approaches explicitly decompose inference into Navigation Visualization and Instruction Generation with Visual Cues, enhancing interpretability and generalization (Fig.~\ref{fig:method overview}(b)).

\noindent\textbf{One-Pass Multimodal Reasoning.}
This approach employs a sequential visual forecasting strategy to predict a complete trajectory of visual observations from the initial state to the goal. Given an initial visual observation sequence $\mathcal{O}_{\mathrm{init}} = \{o_1, \dots, o_k\}$ and a goal observation $o_g$, the model iteratively predicts subsequent visual frames: $ \hat{\mathcal{O}} = \{\hat{o}_{k+1}, \dots, \hat{o}_{k+t}\}$
until a predicted observation $\hat{o}_{k+t}$ satisfies the visual similarity criterion defined by the Structural Similarity Index (SSIM)~\cite{wang2004ssim}:
$ \mathrm{SSIM}(\hat{o}_{k+t}, o_g) > \tau$.
We then strategically select $m{-}1$ representative intermediate frames:$
\{\hat{o}_{i_1}, \dots, \hat{o}_{i_{m-1}}\}$ from the sequence $
\{o_2, \dots, o_k, \hat{o}_{k+1}, \dots, \hat{o}_{k+t}\}$
as inputs to generate the final navigation instruction:
\begin{equation}
\setlength\abovedisplayskip{1pt}
\setlength\belowdisplayskip{1pt}
I = F_\Theta\bigl(\{o_1, \hat{o}_{i_1}, \dots, \hat{o}_{i_{m-1}}, o_g\}\bigr).
\end{equation}
where $i_1, \dots, i_{m-1}$ indicate indices of sampled intermediate frames. This method emphasizes holistic visual context and global scene understanding.

\begin{table*}[!ht]
\belowrulesep=0pt
\aboverulesep=0pt
\renewcommand\arraystretch{1}
\setlength{\abovecaptionskip}{2pt}
\setlength{\belowcaptionskip}{2pt}
\small
\centering
\resizebox{\linewidth}{!}{
\begin{tabular}{l|cccc|cccc|cccc}
\toprule
\multirow{2}{*}{\textbf{Method}} & 
\multicolumn{4}{c|}{\textbf{Validation (Seen)}} & 
\multicolumn{4}{c|}{\textbf{Validation (Unseen)}} & 
\multicolumn{4}{c}{\textbf{Test}} \\
&\textbf{BL-4} $\uparrow$ & \textbf{CI} $\uparrow$& \textbf{ME} $\uparrow$ & \textbf{RO-L} $\uparrow$  
&\textbf{BL-4}  $\uparrow$& \textbf{CI} $\uparrow$& \textbf{ME} $\uparrow$ & \textbf{RO-L} $\uparrow$  
&\textbf{BL-4} $\uparrow$ & \textbf{CI} $\uparrow$& \textbf{ME} $\uparrow$& \textbf{RO-L} $\uparrow$  \\
\midrule
Speaker-Follower \cite{fried2018speakerfollowermodelsvisionandlanguagenavigation}
& 0.10 & 0.08 & 0.08 & 0.12 
& 0.09 & 0.06 & 0.07 & 0.11 
& 0.09 & 0.06 & 0.07 & 0.12 \\

LANA \cite{wang2023lanalanguagecapablenavigatorinstruction}
& 0.05 & 0.05 & 0.11 & 0.10 
& 0.05 & 0.06 & 0.10 & 0.10 
& 0.05 & 0.03 & 0.11 & 0.09 \\

GPT-4o \cite{hurst2024gpt4o}
& 0.08 & 0.16 & \underline{0.19} & 0.18 
& 0.07 & 0.16 & \underline{0.17} & 0.19 
& 0.07 & 0.15 & \underline{0.18} & 0.18 \\

GPT-4o + CoT (zero-shot CoT)
& 0.08 & 0.17 & 0.17 & 0.21 
& 0.09 & 0.16 & \textbf{0.18} & \underline{0.20} 
& 0.08 & 0.17 & \underline{0.18} & 0.19 \\

C-Instructor \cite{Kong_2024}
& 0.21 & 0.19 & 0.14 & \underline{0.23} 
& 0.22 & \underline{0.19} & 0.14 & 0.19
& 0.22 & \underline{0.18} & 0.13 & \underline{0.20} \\

Gemini 2.0~\cite{gemini2blog} 
& 0.08 & 0.11 & 0.13 & 0.12 
& 0.06 & 0.12 & 0.15 & 0.14 
& 0.06 & 0.10 & 0.14 & 0.13 \\

Gemini 3.0~\cite{gemini3}
& 0.09 & 0.13 & 0.16 & 0.13  
& 0.09 & 0.14 & 0.15 & 0.15  
& 0.08 & 0.12 & 0.15 & 0.13  \\

Claude 3 Opus~\cite{anthropic2024claude3} 
& 0.07 & 0.12 & 0.12 & 0.12  
& 0.06 & 0.11 & 0.13 & 0.12  
& 0.07 & 0.11 & 0.13 & 0.13  \\

Claude 4 Opus~\cite{claude4opus} 
& 0.10 & 0.15 & 0.17 & 0.14  
& 0.09 & 0.13 & 0.16 & 0.14  
& 0.09 & 0.14 & 0.17 & 0.14  \\\hdashline

Anole-7B (Direct)~\cite{chern2024anole} 
& 0.06 & 0.10 & 0.10 & 0.14 
& 0.06 & 0.09 & 0.12 & 0.13 
& 0.05 & 0.09 & 0.10 & 0.13 \\

Anole-7B + CoT (fine-tuned CoT)
& 0.10 & 0.14 & 0.15 & 0.17 
& 0.09 & 0.13 & 0.12 & 0.16 
& 0.09 & 0.10 & 0.13 & 0.15 \\

\rowcolor{orange!10}\textbf{Anole-7B + One-pass (Ours)} 
&\underline{0.34} & \underline{0.20} & 0.18 & 0.22 
& \underline{0.29} & 0.18 & 0.16 & \underline{0.20} 
& \underline{0.29} & \textbf{0.19} & 0.17 & 0.18 \\

\rowcolor{blue!10}\textbf{Anole-7B + Interleaved (Ours)}
& \textbf{0.36} & \textbf{0.22} & \textbf{0.21} & \textbf{0.27} 
& \textbf{0.32} & \textbf{0.20} & \textbf{0.18} & \textbf{0.21} 
& \textbf{0.33} & \underline{0.18} & \textbf{0.20} & \textbf{0.22} \\

\bottomrule
\end{tabular}}
\caption{\small \textbf{Comparison with SOTA Methods} on Goal-Conditioned Visual Navigation Instruction Generation on R2R-Goal validation (seen/unseen) and test splits. ({\small BLEU-4 (BL-4), CIDEr (CI), METEOR (ME), and ROUGE-L (RO-L))}}
\label{tab:sota_instruction_comparison}
\vspace{-0.2cm}
\end{table*}

\noindent\textbf{Interleaved Multimodal Reasoning.}~The strategy alternates visualization and instruction generation at each inference step. Initially, the model predicts the immediate next frame $\hat{o}_{k+1}$ based on the init observation sequence $\mathcal{O}_{\mathrm{init}} = \{o_1, \dots, o_k\}$ and goal $o_g$, subsequently generating a preliminary instruction $I_1$ that incorporates updated visual context.~Such an iterative cycle of alternating visual predictions and incremental instruction refinements continues until predicted observation aligns visually with goal ($\mathrm{SSIM}(\hat{o}_{k+t}, o_g) > \tau$). Formally, at inference step $t$, updated instruction $I_t$ is obtained:
\begin{equation}
\setlength\abovedisplayskip{1pt}
\setlength\belowdisplayskip{1pt}
\!I_t\!\!=\!\!F_\Theta(\{\,o_t,\!\dots,\!o_k, \hat{o}_{k+1},\!\dots,\!\hat{o}_{k+t},\!o_g,\!I_{t-1}\}).
\end{equation}
This approach effectively mimics human-like cognitive cycles of visual imagining and linguistic description, enhancing the model’s spatial reasoning and context adaptability. $\tau$ in both reasoning strategies is set to 0.7 in our study. Detailed pseudo-code describing the procedure is provided in Appendix.

\noindent\textbf{Discussion.}
The proposed multimodal reasoning strategies offer two key advantages: (1) Coordinate-Free and Action-Free Reasoning, enabling robust generalization across diverse visual environments without explicit positional or semantic maps; (2) Explicit Visual Reasoning, simulating mental imagery and providing enhanced transparency and interpretability in navigation instruction generation.

\section{Experiments}
\label{sec:experiment}
\subsection{Experimental Setup}
\noindent\textbf{Evaluation Metrics.} We evaluate performance with two sets of metrics:~(1)~\textbf{Instruction Quality.} We measure linguistic accuracy for both the GoViG task and the instruction generation with visual cues subtask using BLEU-4~\cite{papineni2002BLEU}, CIDEr~\cite{vedantam2015cider}, METEOR~\cite{banerjee2005meteor}, and ROUGE-L~\cite{lin2004rouge}. Additionally, we employ navigation-specific metrics Success Rate (SR) and Success weighted by Path Length (SPL)~\cite{anderson2018vision} to assess instruction quality from a practical usability perspective.~(2)~\textbf{Visualization Quality.} We assess visual prediction quality for the navigation visualization subtask using structural and perceptual metrics: SSIM~\cite{wang2004ssim}, PSNR~\cite{hore2010image}, LPIPS~\cite{zhang2018unreasonable}, and DreamSim~\cite{fu2023dreamsim}.

\noindent \textbf{Baselines.}~We benchmark our method against several leading approaches on R2R-Goal: Speaker-Follower~\cite{fried2018speakerfollowermodelsvisionandlanguagenavigation}, LANA~\cite{wang2023lanalanguagecapablenavigatorinstruction}, and C-Instructor~\cite{Kong_2024}, retrained to accept only egocentric observations consistent with our task setting.~We also evaluate SOTA multimodal LLMs including Gemini 2.0~\cite{gemini2blog}, Gemini 3.0~\cite{gemini3}, and Claude 4 Opus~\cite{claude4opus} using direct prompting, as well as GPT-4o~\cite{hurst2024gpt4o} using zero-shot direct and CoT prompting~\cite{wei2022chain,yang2023mm}, and Anole-7B~\cite{chern2024anole} with zero-shot direct prompting and fine-tuned CoT reasoning.

\noindent\textbf{Implementation Details.}~We fine-tune GAIR Anole-7B~\cite{chern2024anole} (4096-token context), freezing both text and image tokenizers. Input images ($256\times256$) are discretized into 784 visual tokens. Only LoRA~\cite{hu2022lora} adapters (rank=16) in the transformer’s \textit{qkv}-projections are updated during training~\cite{liu2023llava}. We train for 20 epochs using AdamW (learning rate=$2\times10^{-4}$). Training employs 4$\times$NVIDIA A100 GPUs (80GB), global batch size 8 (per-GPU batch=1, gradient accumulation=2). Please refer to the Appendix for detailed implementation and metrics.

\begin{table}[!t]
\belowrulesep=0pt
\aboverulesep=0pt
\renewcommand\arraystretch{1.3}
\setlength{\abovecaptionskip}{3pt}
\setlength{\belowcaptionskip}{3pt}
\small
\centering
\resizebox{\linewidth}{!}{
\begin{tabular}{l|cccc}
\toprule
\textbf{Method} & \textbf{SSIM $\uparrow$}& \textbf{PSNR $\uparrow$} & \textbf{LPIPS $\downarrow$} & \textbf{DreamSim $\downarrow$}\\
\midrule
GPT-4o + DALL·E& 0.29 & 9.57 & 0.72 & 0.61 \\
Anole-7B (Direct)& 0.50 & 14.98 & 0.39 & 0.27 \\
\rowcolor{blue!10}\textbf{Ours} & \textbf{0.69} & \textbf{20.02} & \textbf{0.27} & \textbf{0.13} \\
\bottomrule
\end{tabular}}
\caption{\small \textbf{Navigation Visualization Comparison} on R2R-Goal val unseen. Higher SSIM and PSNR, and lower LPIPS and DreamSim reflect superior visual fidelity.}
\label{tab:sota_nav_visual_subtask}
\vspace{-0.1cm}
\end{table}
\begin{table}[!t]
\belowrulesep=0pt
\aboverulesep=0pt
\renewcommand\arraystretch{1.3}
\setlength{\abovecaptionskip}{3pt}
\setlength{\belowcaptionskip}{3pt}
\scriptsize
\centering
\resizebox{\linewidth}{!}{
\begin{tabular}{l|cccc}
\toprule
\textbf{Method} & \textbf{~~SSIM $\uparrow$}& \textbf{~~PSNR $\uparrow$}& \textbf{~~LPIPS $\downarrow$}& \textbf{~~DreamSim $\downarrow$}\\
\midrule
w/o $\mathcal{L}_{vis}$ & 0.52 & 15.35 & 0.36 & 0.23 \\
\rowcolor{blue!10}w/ $\mathcal{L}_{vis}$ & \textbf{0.69} & \textbf{20.02} & \textbf{0.27} & \textbf{0.13} \\
\bottomrule
\end{tabular}}
\caption{\small \textbf{Ablation} of Token Discrepancy Loss ($\mathcal{L}_{vis}$) on navigation visualization (val unseen), context size = 2.}
\label{tab:anole7b_token_discrepancy}
\vspace{-0.2cm}
\end{table}

\subsection{Comparison to SOTA Methods}
\label{sec:sota}
\noindent\textbf{Goal-conditioned Instruction Generation.}~Table~\ref{tab:sota_instruction_comparison} showcases that our proposed One-pass and Interleaved Multimodal Reasoning strategies outperform all baseline methods in GoViG task. 
These results demonstrate that incorporating navigation visualization enhances both contextual grounding and linguistic coherence.~Notably, interleaved reasoning enables progressive integration of fine-grained visual semantics into instruction generation.~Qualitative examples in Fig.~\ref{fig:unseen_result} further illustrate that both methods generate high-quality, visually-grounded instructions on challenging unseen split.

\noindent\textbf{Navigation Visualization.} 
In Table~\ref{tab:sota_nav_visual_subtask}, we compare the performance of our fine-tuned Anole-7B model on the navigation visualization subtask against two baselines: GPT-4o with integrated DALL·E (via the GPT-4o API, where image generation is handled by the DALL·E module) and Anole-7B with direct prompting. The results show that our approach significantly outperforms both baselines across all evaluation metrics.~Notably, our method improves structural fidelity and visual realism (SSIM: 0.69, PSNR: 20.02) in predicted observations. 


\begin{figure}[!t]

\vspace{-0.2cm}
\setlength{\abovecaptionskip}{3pt}
\setlength{\belowcaptionskip}{3pt}
\centering
\includegraphics[width=1\linewidth]{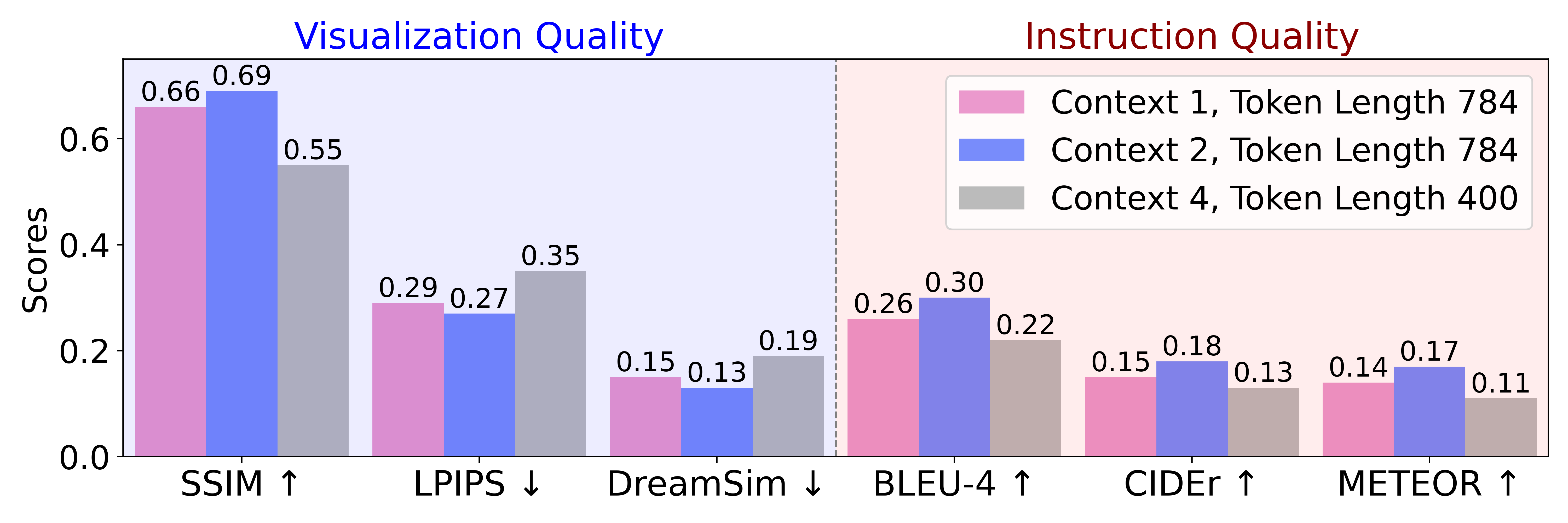}
\caption{\small \textbf{Trade-off} between context size and image token length on both navigation visualization and instruction generation subtasks evaluated on R2R-Goal val unseen split. Context 1 corresponds to context size $k=1$ for visualization and context size $m=2$ for instruction generation. Token Length indicates the number of visual tokens per input or predicted frame.}
\label{fig:content}
\vspace{-0.2cm}
\end{figure}
\subsection{Ablation Studies}\label{sec:ablation}

\noindent\textbf{Impact of Context Size \& Image Token Length.} 
We analyze the influence of context size and image token length in Figure~\ref{fig:content}. Due to Anole-7B's 4096-token constraint, larger contexts (size=4 for visualization, size=5 for instruction generation) necessitate reducing image tokens from 784 to 400 per frame. Results illustrate a clear performance trade-off: moderate context extensions (1→2 frames) enhance temporal coherence and task accuracy, while further expansions at reduced image token length (400 tokens/frame) impair visual fidelity and instruction quality. The results indicate that longer visual histories are effective when each frame retains sufficient token content; otherwise, added context may impede performance.

\noindent\textbf{Effect of Token Discrepancy Loss.}~Table~\ref{tab:anole7b_token_discrepancy} demonstrates the effectiveness of the token discrepancy loss ($\mathcal{L}_{vis}$). Here, w/o $\mathcal{L}_{vis}$ means using label smoothing loss $\mathcal{L}_{ins}$ instead on image tokens. $\mathcal{L}_{vis}$ substantially improves image quality across all metrics, indicating that explicitly modeling token similarity is crucial for preserving perceptual and structural details in visual predictions.


\noindent \textbf{Computational Efficiency Trade-off.}~We analyze the computational cost of our two reasoning strategies on the R2R-Goal validation unseen split. One-pass reasoning achieves 1.2$\times$ faster inference than Interleaved reasoning. This difference arises from Interleaved's iterative instruction refinement at each step.~Besides, this presents a practical trade-off: Interleaved delivers higher accuracy (BLEU-4: 0.32 vs 0.29) with a 20\% time cost, suitable for those quality-critical applications, while One-pass offers efficient inference when speed is prioritized.


\subsection{Instruction Quality Analysis}
While n-gram-based metrics (BLEU-4, CIDEr) provide quantitative assessments of instruction quality, they may not fully capture instruction usefulness or human interpretability. We therefore conduct additional analyzes to evaluate our generated instructions from two complementary perspectives:

\noindent \textbf{Practical Usability.}~The navigation performance of agents following instructions from different generators serves as an indicator of instruction quality. We regenerate instructions for paths in the R2R-Goal val unseen split and employ two navigators (ETPNav and BEVBert) to evaluate SR and SPL under the regenerated guidance. As shown in Table~\ref{tab:usability}, instructions generated by our methods (one-pass and interleaved) achieve competitive results that exceed those of prior models and closely align with the navigation accuracy obtained using human-annotated instructions.

\begin{figure*}[!t]
\setlength{\abovecaptionskip}{3pt}
\setlength{\belowcaptionskip}{3pt}
\centering
    \includegraphics[width=0.98\textwidth]{ 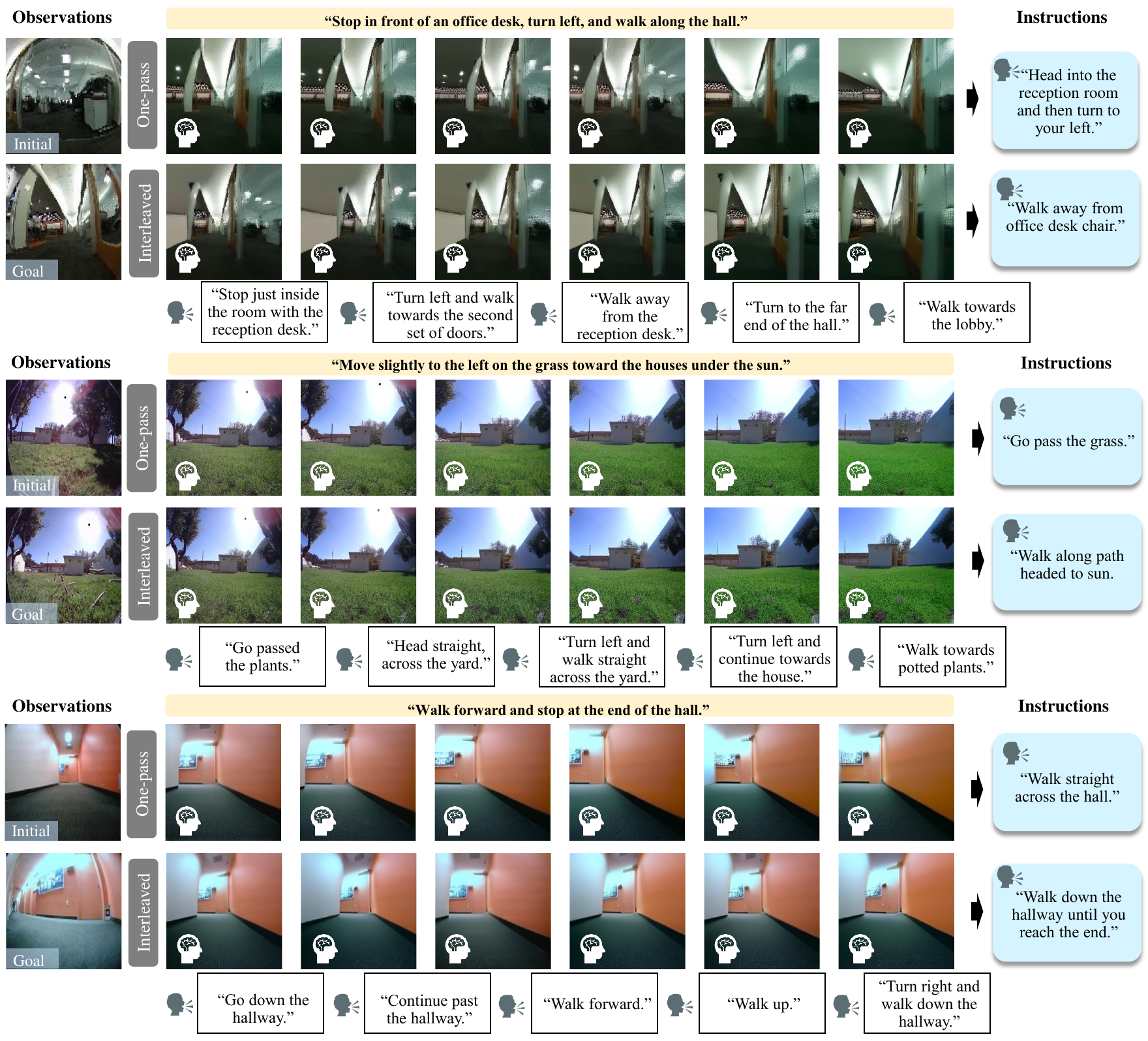}
\caption{\small \textbf{Qualitative results} on real-world subset of (with ground truth) our One-pass and Interleaved Reasoning.}
\label{fig:app_social}
\vspace{-0.2cm}
\end{figure*}

\begin{table}[t]
\belowrulesep=0pt
\aboverulesep=0pt
\renewcommand\arraystretch{1.1}
\setlength{\abovecaptionskip}{2pt}
\setlength{\belowcaptionskip}{2pt}
\scriptsize
    \centering
    \resizebox{\linewidth}{!}{
    \begin{tabular}{l|p{1.1cm}<{\centering}p{1.1cm}<{\centering}|p{1.1cm}<{\centering}p{1.1cm}<{\centering}}
    \toprule[1pt]
         \multirow{2}{*}{\textbf{Instruction Generator}}& \multicolumn{2}{c|}{\textbf{ETPNav \cite{an2024etpnav}}} & \multicolumn{2}{c}{\textbf{BEVBert \cite{an2023bevbert}}} \\\cline{2-5}
         & \textbf{SR $\uparrow$}&\textbf{SPL $\uparrow$}&\textbf{SR $\uparrow$}&\textbf{SPL $\uparrow$}\\\hline
         \textit{Human Annotation} &0.36 &0.28 &0.34 &0.27 \\\hdashline
         LANA&0.18 &0.11 &0.17 &0.12 \\ 
         
         GPT-4o&0.23 &0.16 &0.22 &0.14 \\ 
         
         GPT-4o + CoT&0.25 &0.17 &0.24 &0.17 \\ 
         
         C-Instructor&0.29 &0.19 &0.27 &0.18 \\
         
         Gemini 3.0&0.27 &0.18 &0.25 &0.14 \\ 
         
         Claude 4 Opus &0.26 &0.16 &0.25 &0.17 \\
         
         Anole-7B + Direct &0.20 &0.14 &0.18 &0.13 \\ 
         
         Anole-7B + CoT&0.25 &0.16 &0.23 &0.15 \\
         \rowcolor{orange!10}\textbf{Anole-7B + One-pass}&\underline{0.31} &\underline{0.20} &\underline{0.29} &\underline{0.21}\\
         \rowcolor{blue!10}\textbf{Anole-7B + Interleaved}&\textbf{0.34} &\textbf{0.25}  &\textbf{0.33} &\textbf{0.25}\\
    \bottomrule[1pt]
    \end{tabular}}
    \caption{\small \textbf{Instruction Quality Analysis.} Performance of ETPNav~\cite{an2024etpnav} and BEVBert~\cite{an2023bevbert} in following instructions generated on R2R-Goal val unseen.}
    \label{tab:usability}
    \vspace{-0.1cm}
\end{table}
\begin{table}[!t]
\belowrulesep=0pt
\aboverulesep=0pt
\renewcommand\arraystretch{1.1}
\setlength{\abovecaptionskip}{2pt}
\setlength{\belowcaptionskip}{2pt}
\small
\setlength{\tabcolsep}{2pt}
\resizebox{\linewidth}{!}{
\centering
\begin{tabular}{l|cccc}
\toprule[1.2pt]
\textbf{Method}& \textbf{BLEU-4 $\uparrow$}& \textbf{CIDEr $\uparrow$}& \textbf{METEOR $\uparrow$}& \textbf{ROUGE-L $\uparrow$}\\
\midrule
LANA& 0.05 & 0.03 & 0.09 & 0.09 \\
GPT-4o& 0.08 & 0.11 & \underline{0.18} & 0.17 \\
GPT-4o + CoT & 0.09 & 0.13 & 0.16 & 0.18 \\
C-Instructor& 0.15 & 0.08 & 0.12 & 0.15 \\
Gemini 3.0 & 0.08 & 0.11 & 0.15 & 0.14 \\
Claude 4 Opus & 0.09 & 0.13 & 0.16 & 0.16 \\
Anole-7B + Direct & 0.06 & 0.09 & 0.10 & 0.12 \\
Anole-7B + CoT & 0.08 & 0.10 & 0.13 & 0.17 \\
\rowcolor{orange!10}\textbf{Anole-7B + One-pass}& \underline{0.24} & \underline{0.14} & 0.17 & \underline{0.19} \\
\rowcolor{blue!10}\textbf{Anole-7B + Interleaved}& \textbf{0.27} & \textbf{0.15} & \textbf{0.19} & \textbf{0.20} \\
\bottomrule[1.2pt]
\end{tabular}}
\caption{\small \textbf{Zero-shot generalization} on R2R-Goal real-world subset (GO Stanford, ReCon, and HuRoN). All models here are evaluated without fine-tuning on this subset.}
\label{tab:sota_cross-domain_instruction_comparison}
    \vspace{-0.2cm}
\end{table}

\noindent \textbf{User Study.}~To evaluate instruction quality beyond automatic metrics, we recruit 21 anonymous participants from diverse backgrounds to score instructions from 1 to 6 based on semantic alignment with trajectories. We evaluate our two reasoning strategies and SOTA methods on 320 randomly sampled trajectories from R2R-Goal val unseen split, presented in randomized order. Our Interleaved reasoning achieves the highest score of 4.85, followed by One-pass at 4.52, outperforming GPT-4o + CoT (3.76), Gemini 3.0 (3.54), Claude 4 Opus (3.41), C-Instructor (3.08), and LANA (2.67).

\subsection{Cross-Domain Generalization}\label{sec:cross-domain}
To comprehensively assess cross-domain generalization, we evaluate our method on the real-world subset of R2R-Goal, comprising diverse scenes from GO Stanford, ReCon, and HuRoN. As shown in Table~\ref{tab:sota_cross-domain_instruction_comparison}, our Interleaved and One-Pass multimodal reasoning strategies notably outperform SOTA approaches under significant domain shifts. Specifically, the Interleaved strategy consistently delivers superior results, underscoring the efficacy of iterative visual-linguistic refinement in improving contextual grounding and instruction coherence. Qualitative examples (Fig.~\ref{fig:app_social}) further illustrate how iterative multimodal reasoning mirrors adaptive human cognition, enabling robust instruction generation even in challenging, unseen environments.


\vspace{-0.1cm}

\section{Conclusion}
In this work, we proposed Goal-Conditioned Visual Navigation Instruction Generation (GoViG), a framework to generate precise and context-aware navigation instructions solely from egocentric visual observations, eliminating reliance on privileged data such as maps or semantic annotations. Our approach systematically integrates two interdependent subtasks, Navigation Visualization and Instruction Generation with Visual Cues, into a unified autoregressive MLLM. Furthermore, we developed two multimodal reasoning strategies (One-Pass and Interleaved) that enhance spatial reasoning and linguistic coherence. Comprehensive experiments on our proposed R2R-Goal benchmark demonstrate superior instruction quality and robust cross-domain generalization. Future directions include exploring real-time environmental feedback to advance practical embodied AI.

\section*{Acknowledgments}

This work was supported by the University of Washington Faculty Startup Fund, the Carwein Andrews Fellowship, the UW GSFEI Top Scholar Award, and the U.S. DOT PacTrans sub-center seed funding program. We thank the anonymous reviewers for their helpful comments.

\appendix

\section*{Appendix}
\noindent This supplementary material provides expanded details and results that complement the main paper. Section~\ref{appx:sec-more-related-work} presents a more detailed analysis of related work and comparisons of navigation instruction generation methods. Section~\ref{appx-sec:method-details} includes details on the R2R-Goal dataset, pseudo-code for the one-pass and interleaved multimodal reasoning mechanisms, as well as prompt design specifications. Section~\ref{appx-sec:exp-details} reports detailed implementation specifics and additional qualitative results.

\begin{table*}[!h]
\belowrulesep=0pt
\aboverulesep=0pt
\renewcommand\arraystretch{1.1}
\setlength{\abovecaptionskip}{1pt}
\setlength{\belowcaptionskip}{0pt}
\centering
\small
\setlength{\tabcolsep}{2pt}
\resizebox{\linewidth}{!}{
\begin{tabular}{
>{\centering\arraybackslash}p{2.6cm}|
>{\centering\arraybackslash}c|
>{\centering\arraybackslash}p{3.6cm}|
>{\centering\arraybackslash}p{3.8cm}|
>{\centering\arraybackslash}c|
>{\centering\arraybackslash}p{2cm}
}
\toprule
\textbf{Method} & \textbf{Ego-centric} & \textbf{Privileged Input} & \textbf{Pre-processed Elements} & \textbf{Backbone} & \textbf{LLM Usage} \\
\midrule
Speaker-Follower \cite{fried2018speakerfollowermodelsvisionandlanguagenavigation} & \ding{53} & Panoramic views, Action history & ResNet features, GloVe embeddings & LSTM-RNN & None \\

CCC-Speaker \cite{9879110} & \ding{53} & Panoramic views, Action, Environment labels & ResNet features & CNN + LSTM & None \\

LANA \cite{wang2023lanalanguagecapablenavigatorinstruction} & \ding{53} & Panoramic views, Orientation & - & Transformer & None \\

LANA+ \cite{10359152} & \ding{53} & Panoramic views, Orientation & Landmark spotting via CLIP & Transformer + CLIP & None \\

C-Instructor \cite{Kong_2024} & \ding{53} & Trajectory path, Panoramic views, Action, Orientation & Landmark vocabulary (CoTL)& Vision Encoder + LLM & Chain-of-Thought \\

BEV-Instructor \cite{fan2024navigation} & \ding{51}\rotatebox[origin=c]{-9.2}{\kern-0.7em\ding{55}} & Multi-view images, Orientation, Action, 3D bounding boxes & BEV encoding, Action map & Vision Encoder + MLLM & Yes \\

NavRAG \cite{wang2025navraggeneratinguserdemand} & \ding{53} & Navigable position, Panoramic views, GPS & Hierarchical scene tree & Vision Encoder + LLM & Retrieval-Augmented Generation \\

MapInstructor \cite{Fan_2025_CVPR} & \ding{53} & Panoramic views, Action, Orientation & Landmark extraction via scene map & CLIP + GCN + LLM & Map-based Prompt Tuning \\

\textbf{Ours} & \ding{51} & Not Required & Not Required & Anole-7B & Multi-modal Reasoning \\
\bottomrule
\end{tabular}}
\caption{\small Comparison of navigation instruction generation methods. Abbreviations: CoTL = Chain-of-Thought with Landmarks, BEV = Bird’s Eye View, GCN = Graph Convolutional Network, MLLM = Multi-modal Large Language Model.}
\label{tab:detail}
\vspace{-0.3cm}
\end{table*}

\section{More Related Work}

\label{appx:sec-more-related-work}
Table~\ref{tab:detail} categorizes prior work along five orthogonal axes: (i) viewpoint (\emph{ego-centric} vs. \emph{panoramic}); (ii) reliance on privileged inputs (e.g., orientation, GPS, environment labels); (iii) pre-processing pipelines (e.g., landmark vocabularies, BEV encodings, scene graphs); (iv) backbone family (RNN/CNN, Transformer, CLIP/GCN, Vision-Encoder + LLM); and (v) the extent and manner in which LLMs are incorporated.
Early “speaker-style” systems, Speaker-Follower~\cite{fried2018speakerfollowermodelsvisionandlanguagenavigation} and CCC-Speaker~\cite{9879110}, adopt a non-ego-centric, panoramic observation paradigm with action traces, occasionally augmented by environment labels. These methods typically depend on pre-extracted visual and linguistic features (e.g., ResNet, GloVe) and sequence backbones (CNN/LSTM), without leveraging any large language models.
Transformer-based approaches, such as LANA~\cite{wang2023lanalanguagecapablenavigatorinstruction} and LANA+~\cite{10359152}, retain the panoramic setting but incorporate orientation priors and stronger sequence modeling. LANA+ further introduces CLIP-based landmark spotting as an explicit pre-processing signal, improving visual grounding while still assuming privileged panoramic inputs.

Recently, LLM-integrated “instructor” approaches have broadened the modeling toolkit but often at the cost of introducing stronger priors and heavier pre-processing pipelines. C-Instructor~\cite{Kong_2024} couples a vision encoder with an LLM and a curated landmark vocabulary, employing Chain-of-Thought prompting to scaffold instruction generation. BEV-Instructor~\cite{fan2024navigation} moves toward an ego-centric perspective but still depends on multi-view imagery, 3D bounding boxes, and BEV/action-map encodings orchestrated by an MLLM. Retrieval- and map-centric variants, NavRAG~\cite{wang2025navraggeneratinguserdemand} and MapInstructor~\cite{Fan_2025_CVPR}, leverage navigable positions, panoramic imagery, GPS, and scene maps to construct hierarchical structures or extract landmarks, then condition an LLM via RAG or map-based prompt tuning.

In contrast, our method operates \emph{exclusively} on ego-centric inputs, free of privileged priors or handcrafted pre-processing. By harnessing the Multimodal LLM \emph{Anole-7B} for unified multi-modal reasoning, it intentionally minimizes task-specific engineering~(e.g.,~curated vocabularies,~BEV~encodings,~retrieval indices) yet preserves strong grounding performance.~Our~design facilitates practical deployment and promotes robust cross-domain generalization by eliminating dependencies on panoramic sensors, external maps, or GPS signals.

\section{Methodology Details}
\label{appx-sec:method-details}

\subsection{R2R-Goal Dataset Details}
To support the GoViG task, we construct the R2R-Goal dataset within the HA-VLN simulation environment~\cite{dong2025ha}, using the path start and goal positions provided by the HA-R2R~\cite{dong2025ha} and R2R-CE~\cite{krantz_vlnce_2020} benchmarks. An A*-based heuristic search identifies the shortest feasible navigation path, with dynamic re-planning triggered in real time upon encountering unexpected obstacles. An egocentric camera mounted on the simulated agent continuously captures observations along each traversed path.
Scene-level segmentation is performed in two stages using a frozen Qwen2.5-VL-7B-Instruct model~\cite{bai2025qwen2}. First, navigation instructions are segmented into spatially coherent scenes, ensuring each segment corresponds to a navigable space and that all text is uniquely assigned. Post-processing merges consecutive identical scenes and guarantees complete coverage, yielding scene–instruction pairs (e.g., “Kitchen” as an instruction segment). Second, observation frames are aligned with the segmented scenes: the model analyzes the full visual sequence to detect scene transitions based on visual cues and instruction alignment, followed by post-processing to adjust boundaries and eliminate gaps or overlaps.

Annotators were instructed to align their navigation descriptions closely with the visual scenes presented, ensuring that the language reflected the perspective shifts between initial and goal viewpoints. They were encouraged to use diverse expressions when describing actions, environments, and spatial relations among objects, including references to relative positions (e.g., left/right, near/far), motion dynamics (e.g., slow approach, rapid turn), and changes in viewpoint. This emphasis on alignment and variation was intended to capture richer correspondences between instructions and observations, while avoiding repetitive phrasing.

\begin{figure*}
    \centering
    \includegraphics[width=1\linewidth]{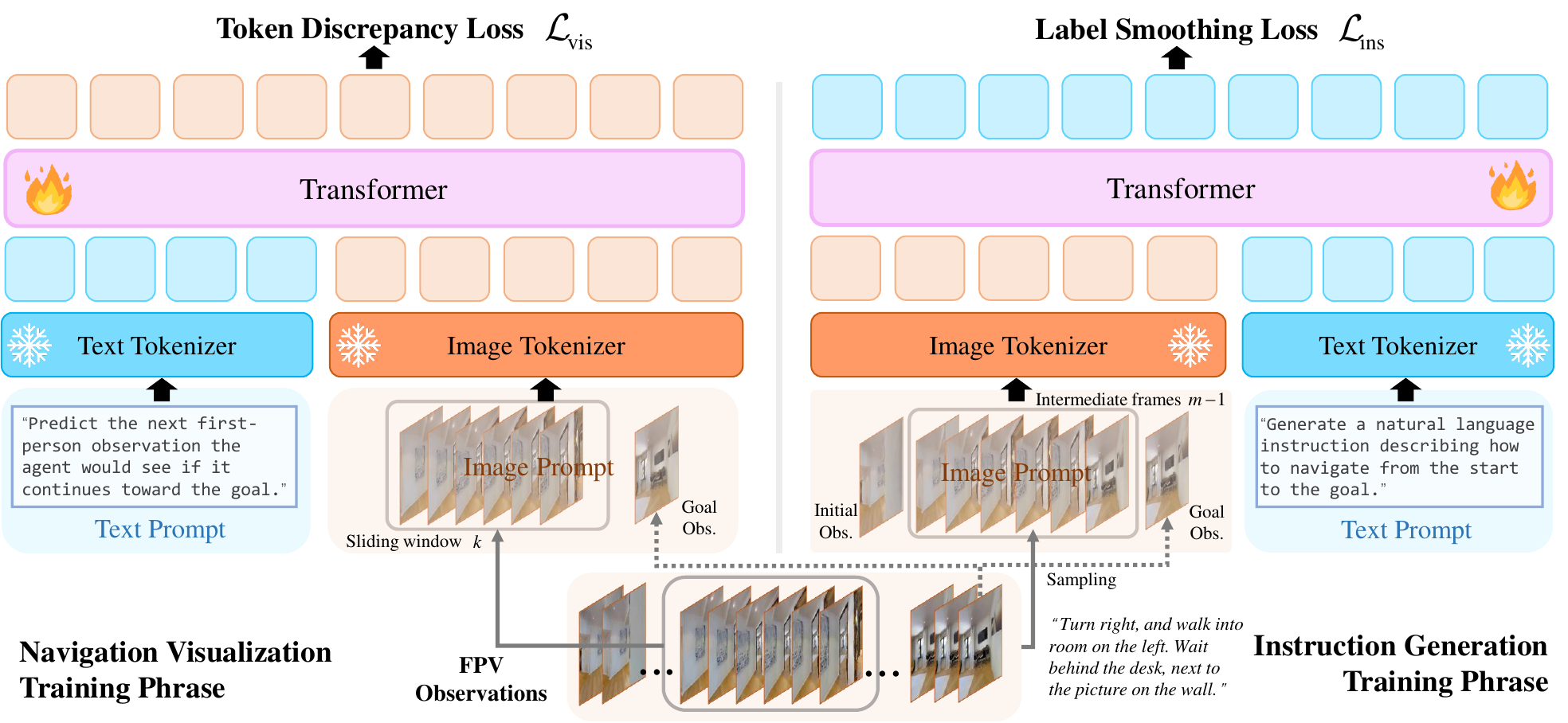}
    \caption{Detailed structure and pipeline of our training procedure.}
    \label{fig:detail_training}
\end{figure*}

All datasets used in this work (R2R-CE, HA-R2R, GO Stanford, ReCon, HuRoN) are publicly available research datasets released under appropriate licenses. We did not collect any personal data. During dataset construction, we checked that the included visual and textual data do not contain names, faces, or other information that could uniquely identify individuals. Offensive or harmful content was not observed in the sources used. For the real-world subset, only publicly released egocentric videos were used, and all annotations were created to describe navigation trajectories without reference to personal identity. Thus, no anonymization beyond the original dataset release was required.

The annotators were volunteers recruited through university mailing lists and research group announcements. Participation was entirely voluntary, with no monetary compensation provided. Volunteers contributed time to support academic research, and their involvement was recognized in accordance with established ethical guidelines.

\subsection{Autoregressive MLLM Training Details}
Fig.~\ref{fig:detail_training} illustrates the dual training paradigm of our autoregressive multimodal Transformer, which unifies visual and linguistic reasoning based on the Chameleon architecture~\cite{team2024chameleon}. The left panel depicts the \textbf{Navigation Visualization Training Phase}, where a structured text prompt (e.g., ``Predict the next first-person observation the agent would see if it continues toward the goal.'') is tokenized via a BPE tokenizer, and paired with a sequence of $k$ preceding FPV frames plus the goal frame. These visual observations are encoded as \texttt{<image>} tokens using a vector-quantized (VQ) image tokenizer. The resulting multimodal prompt is processed by a causal Transformer to predict the next visual token, optimized via the Token Discrepancy Loss $\mathcal{L}_{\text{vis}}$, which compares predicted token distributions $P(t_i)$ against ground-truth embeddings $\text{emb}_i$ over the visual codebook $\mathcal{C}$.

The right panel shows the \textbf{Instruction Generation Training Phase}, where FPV frames from the initial to goal observation (including $m-1$ intermediate frames) are encoded into an image prompt. This is paired with a text prompt such as ``Generate a natural language instruction describing how to navigate from the start to the goal,'' and the ground-truth instruction (e.g., ``Turn right, and walk into room on the left. Wait behind the desk, next to the picture on the wall.''). Both image and text inputs are tokenized and fed into the Transformer, which autoregressively generates the instruction sequence. Training is guided by the label smoothing cross-entropy loss $\mathcal{L}_{\text{ins}}$, computed over the vocabulary $\mathcal{V}$ with smoothed targets $q_v(y_i)$ and predicted probabilities $P_v(y_i)$. Samples from both phases are interleaved during training, enabling joint optimization of visual forecasting and instruction generation within a unified multimodal framework.

\begin{algorithm}[!t]
\caption{One-Pass Multimodal Reasoning}
\label{alg:one_pass}
\begin{algorithmic}
\REQUIRE Initial observations $\mathcal{O}_{\mathrm{init}} = \{o_1, \dots, o_k\}$ with visualization context size $k$; goal observation $o_g$; SSIM threshold $\tau$; MLLM $F_\Theta$ with parameters and tokenizers frozen; instruction context size $m$
\ENSURE Final instruction $I$
\STATE Initialize step $t \leftarrow 1$
\STATE Initialize observation context window $\!\hat{\mathcal{O}}^{(t)} \leftarrow \mathcal{O}_{\mathrm{init}}$

\STATE $m= k+1$

\REPEAT
    \STATE $\hat{o}_{k+t} \leftarrow F_\Theta(\hat{\mathcal{O}}^{(t)}, o_g)$
    \STATE $\hat{\mathcal{O}}^{(t+1)} \leftarrow \hat{\mathcal{O}}^{(t)}[2{:}] \cup \{\hat{o}_{k+t}\}$
   
    \COMMENT{Update $\hat{\mathcal{O}}^{(t)}$ by sliding in $\hat{o}_{k+t}$ and keeping most recent $k$ observations}
    \STATE $t \leftarrow t + 1$
\UNTIL{$\mathrm{SSIM}(\hat{o}_{k+t}, o_g) > \tau$}
\STATE Sample $m{-}1$ intermediate frames $\{\hat{o}_{i_1}, \dots, \hat{o}_{i_{m-1}}\}$ from $\{o_2, \dots, o_k, \hat{o}_{k+1}, \dots, \hat{o}_{k+t}\}$
\STATE $I \leftarrow F_\Theta(\{o_1, \hat{o}_{i_1}, \dots, \hat{o}_{i_{m-1}}, o_g\})$
\RETURN $I$
\end{algorithmic}
\end{algorithm}

\begin{algorithm}[!t]
\caption{Interleaved Multimodal Reasoning}
\label{alg:interleaved}
\begin{algorithmic}
\REQUIRE Initial observations $\mathcal{O}_{\mathrm{init}} = \{o_1, \dots, o_k\}$ with visualization context size $k$; goal observation $o_g$; SSIM threshold $\tau$; MLLM $F_\Theta$ with parameters and tokenizers frozen; instruction context size $m$
\ENSURE Final instruction $I$

\STATE Initialize step $t \leftarrow 1$
\STATE Initialize observation context window $\hat{\mathcal{O}}^{(t)} \leftarrow \mathcal{O}_{\mathrm{init}}$ \COMMENT{Initial context window}
\STATE Initialize instruction $I_0 \leftarrow$ empty string
\STATE $m=k+1$
\REPEAT
    \STATE $\hat{o}_{k+t} \leftarrow F_\Theta(\hat{\mathcal{O}}^{(t)}, o_g)$ \COMMENT{Predict next observation}
    \STATE Update $\hat{\mathcal{O}}^{(t+1)} \leftarrow \hat{\mathcal{O}}^{(t)}[2{:}] \cup \{\hat{o}_{k+t}\}$ \COMMENT{Slide in $\hat{o}_{k+t}$}
    \STATE $I_t \leftarrow F_\Theta(\hat{\mathcal{O}}^{(t+1)} \cup \{o_g, I_{t-1}\})$ \COMMENT{Update instruction}
    \STATE $t \leftarrow t + 1$
\UNTIL{$\mathrm{SSIM}(\hat{o}_{k+t}, o_g) > \tau$}
\STATE $I = I_t$
\RETURN $I$
\end{algorithmic}
\end{algorithm}

\subsection{Pseudo-Code of Reasoning Strategies}

To generate instructions from an egocentric initial observation and a goal observation, we propose two multimodal reasoning strategies: \emph{One-Pass} and \emph{Interleaved} reasoning, with their pseudo-code provided in Algorithms~\ref{alg:one_pass} and~\ref{alg:interleaved}, respectively. Both strategies employ a frozen multimodal language model $F_\Theta$ and iteratively visualize navigation until the predicted frame achieves sufficient visual similarity to the goal observation, measured by an SSIM threshold $\tau$.

\noindent\textbf{One-Pass Multimodal Reasoning} (Algorithm~\ref{alg:one_pass}) first generates the entire future trajectory $\hat{\mathcal{O}} = \{\hat{o}_{k+1}, \dots, \hat{o}_{k+t}\}$ using a sliding, fixed-size context window, terminating when $\mathrm{SSIM}(\hat{o}_{k+t}, o_g) > \tau$. It then samples $m{-}1$ representative intermediate frames and produces a final instruction via:
\begin{equation}
I = F_\Theta\big(\{o_1, \hat{o}_{i_1}, \dots, \hat{o}_{i_{m-1}}, o_g\}\big).
\end{equation}

\noindent\textbf{Interleaved Multimodal Reasoning} (Algorithm~\ref{alg:interleaved}) alternates between predicting the next visual frame and incrementally refining the instruction. At each step $t$, the instruction is updated as:
\begin{equation}
I_t = F_\Theta\big(\hat{\mathcal{O}}^{(t+1)} \cup \{o_g, I_{t-1}\}\big),
\end{equation}
continuing until the SSIM criterion is met. This step-wise refinement allows the agent to progressively incorporate new visual cues, potentially improving instruction grounding in dynamically evolving environments.

\subsection{Prompt Design Details and Examples}
We examine the detailed prompt formulation and response behaviors of two multimodal reasoning strategies—\emph{One-Pass} and \emph{Interleaved}—across two navigation subtasks: \emph{Navigation Visualization} and \emph{Instruction Generation with Visual Cues}. These examples illustrate how multimodal inputs guide both visual prediction and instruction generation in visually grounded navigation.

\begin{figure}[!t]
\setlength{\abovecaptionskip}{-3pt}
\setlength{\belowcaptionskip}{0pt}
\newtcolorbox{fullpromptbox}[2]{
  enhanced,
  colback=#1!5,
  colframe=#1!50,
  fonttitle=\bfseries,
  title=#2,
  sharp corners,
  boxrule=0.8pt,
  width=\linewidth,
  lower separated=false,
  bottomtitle=0mm
}
\begin{fullpromptbox}{orange}{\small Navigation Visualization (One-pass)}
\textbf{\tcbox[colback=gray!20, colframe=gray!20, boxrule=0.1pt,top=0mm, bottom=0mm]{~~~~~~~~~~~~~~~~~~~~~~~~~~~~~~~~~~~~~~~~Input~~~~~~~~~~~~~~~~~~~~~~~~~~~~~~~~~~~~~~~~}}

\textbf{Task:} Navigation Single Step Visualization

\textbf{Description:} Given the previous first-person observation, the current first-person observation, and the goal observation, predict the next first-person observation the agent would see if it continues toward the goal.

\textbf{Input obs:}

Previous obs:
~~~~~\includegraphics[height=1.7cm]{ 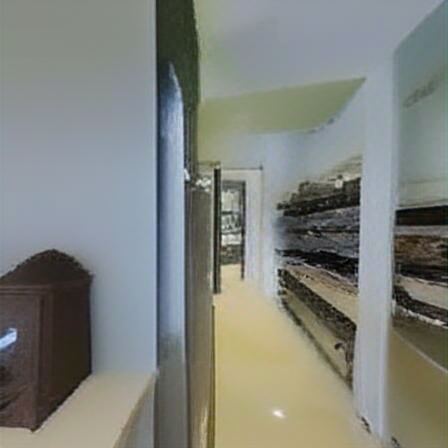}

Current obs:
~~~~~\,~\includegraphics[height=1.7cm]{ 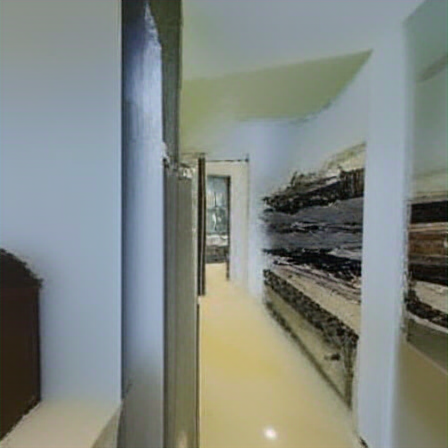}

Goal obs:
~~~~~~~~~~~\includegraphics[height=1.7cm]{ 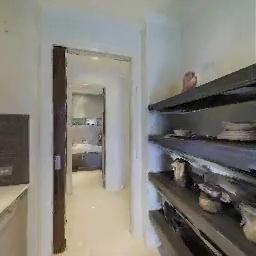}

\textbf{\tcbox[colback=gray!20, colframe=gray!20, boxrule=0.1pt,top=0mm, bottom=0mm]{~~~~~~~~~~~~~~~~~~~~~~~~~~~~~~~~~~~~Response~~~~~~~~~~~~~~~~~~~~~~~~~~~~~~~~~}}
\textbf{Predicted obs:}~\includegraphics[height=1.7cm]{ 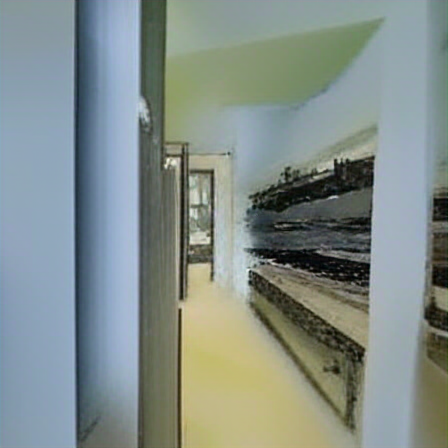}
\end{fullpromptbox}
\caption{\small Prompt design examples on One-pass multimodal reasoning during the inference stage for the Navigation Visualization subtask. (Context size $k$ = 2)}
\label{fig:app_op_nv}
\vspace{-0.2cm}
\end{figure}

\begin{figure}[!t]
\setlength{\abovecaptionskip}{-3pt}
\setlength{\belowcaptionskip}{0pt}
\newtcolorbox{fullpromptbox}[2]{
  enhanced,
  colback=#1!5,
  colframe=#1!50,
  fonttitle=\bfseries,
  title=#2,
  sharp corners,
  boxrule=0.8pt,
  width=\linewidth,
  lower separated=false,
  bottomtitle=0mm
}
\begin{fullpromptbox}{cyan}{\small Instruction Generation (One-pass)}
\textbf{\tcbox[colback=gray!20, colframe=gray!20, boxrule=0.1pt,top=0mm, bottom=0mm]{~~~~~~~~~~~~~~~~~~~~~~~~~~~~~~~~~~~~~~~~Input~~~~~~~~~~~~~~~~~~~~~~~~~~~~~~~~~~~~~~~~}}
\textbf{Task:} Scene-level Instruction Generation

\textbf{Description:} Given a sequence of sampled first-person observations along a navigation trajectory, including the starting observation, several intermediate observations, and the goal observation, generate a natural language instruction describing how to navigate from the start to the goal. Focus on essential actions and prominent, easily identifiable landmarks.

\textbf{Input obs:}

Starting obs:
~~~~~~~~~\includegraphics[height=1.7cm]{ 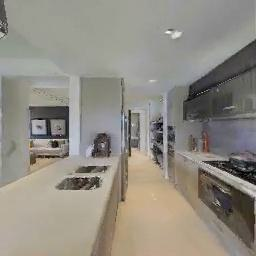}

Intermediate obs: 
~\includegraphics[height=1.7cm]{ 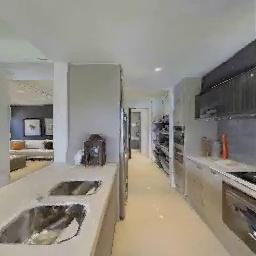}~\includegraphics[height=1.7cm]{ 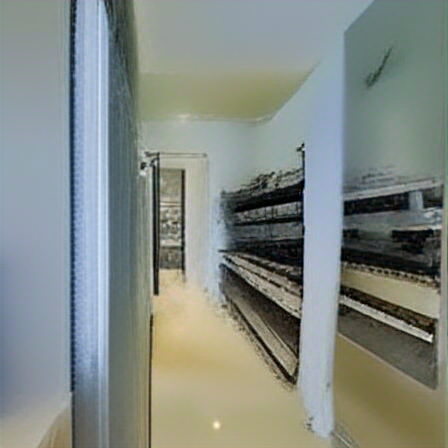}

Goal obs:
~~~~~~~~~~~~~~\includegraphics[height=1.7cm]{img/goal_obs.png}

\textbf{\tcbox[colback=gray!20, colframe=gray!20, boxrule=0.1pt,top=0mm, bottom=0mm]{~~~~~~~~~~~~~~~~~~~~~~~~~~~~~~~~~~~~Response~~~~~~~~~~~~~~~~~~~~~~~~~~~~~~~~~}}
\textbf{Predicted Instruction:}

Walk out of the kitchen.
\end{fullpromptbox}
\caption{\small Prompt design examples on One-pass multimodal reasoning during inference stage for Instruction Generation with Visual Cues. (Context size $m$ = 3)}
\label{fig:app_op_ig}
\vspace{-0.2cm}
\end{figure}

\noindent\textbf{One-Pass Multimodal Reasoning.} As shown in Fig.~\ref{fig:app_op_nv}, the model begins with an initial observation and iteratively predicts future frames toward the goal, updating the context with each new prediction until the generated frame satisfies the SSIM threshold relative to the goal observation. For instruction generation (Fig.~\ref{fig:app_op_ig}), once visual prediction is complete, the model samples key frames—initial, intermediate, and goal—and produces a concise instruction (e.g., “Walk out of the kitchen”) summarizing the visual trajectory.

\begin{figure}[!t]
\setlength{\abovecaptionskip}{-3pt}
\setlength{\belowcaptionskip}{0pt}
\newtcolorbox{fullpromptbox}[2]{
  enhanced,
  colback=#1!5,
  colframe=#1!50,
  fonttitle=\bfseries,
  title=#2,
  sharp corners,
  boxrule=0.8pt,
  width=\linewidth,
  lower separated=false,
  bottomtitle=0mm, 
}
\begin{fullpromptbox}{orange}{\small Navigation Visualization (Interleaved)}
\textbf{\tcbox[colback=gray!20, colframe=gray!20, boxrule=0.1pt,top=0mm, bottom=0mm]{~~~~~~~~~~~~~~~~~~~~~~~~~~~~~~~~~~~~~~~~Input~~~~~~~~~~~~~~~~~~~~~~~~~~~~~~~~~~~~~~~~}}

\textbf{Task:} Navigation Single Step Visualization

\textbf{Description:} Given the previous first-person observation, the current first-person observation, and the goal observation, predict the next first-person observation the agent would see if it continues toward the goal.

\textbf{Input obs:}

Previous obs:
~~~~~~\includegraphics[height=1.7cm]{ img/step_4_obs.png}

Current obs:
~~~~~~\,~\includegraphics[height=1.7cm]{ 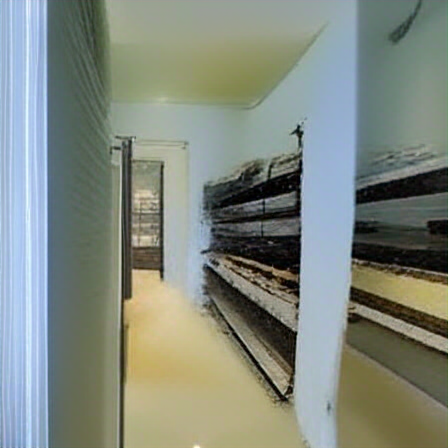}

Goal obs:
~~~~~~~~~~~~\includegraphics[height=1.7cm]{ img/goal_obs.png}

\textbf{\tcbox[colback=gray!20, colframe=gray!20, boxrule=0.1pt,top=0mm, bottom=0mm]{~~~~~~~~~~~~~~~~~~~~~~~~~~~~~~~~~~~~Response~~~~~~~~~~~~~~~~~~~~~~~~~~~~~~~~~}}
\textbf{Predicted observation:}
~\includegraphics[height=1.7cm]{ 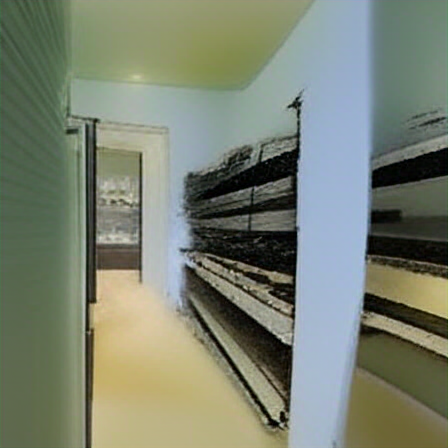}
\end{fullpromptbox}
\caption{\small Prompt design examples on Interleaved multimodal reasoning during the inference stage for the Navigation Visualization subtask. (Context size $k$ = 2)}
\label{fig:app_iter_nv}
\vspace{-0.2cm}
\end{figure}
\begin{figure}[!t]
\setlength{\abovecaptionskip}{-3pt}
\setlength{\belowcaptionskip}{0pt}
\newtcolorbox{fullpromptbox}[2]{
  enhanced,
  colback=#1!5,
  colframe=#1!50,
  fonttitle=\bfseries,
  title=#2,
  sharp corners,
  boxrule=0.8pt,
  width=\linewidth,
  lower separated=false,
  bottomtitle=0mm
}
\begin{fullpromptbox}{cyan}{\small Instruction Generation (Interleaved)}
\textbf{\tcbox[colback=gray!20, colframe=gray!20, boxrule=0.1pt,top=0mm, bottom=0mm]{~~~~~~~~~~~~~~~~~~~~~~~~~~~~~~~~~~~~~~~~Input~~~~~~~~~~~~~~~~~~~~~~~~~~~~~~~~~~~~~~~~}}
\textbf{Task:} Scene-level Instruction Generation

\textbf{Description:} Given a sequence of sampled first-person observations and a previously generated instruction that describes the navigation path — including: the previous instruction (which was originally generated to guide navigation from the start observation to the goal), the current observation (resulting from navigating one step from the previous observation toward the goal), the past observations, and the goal observation. Determine whether the previous instruction needs refinement based on the visual context. Then, generate an updated natural language instruction that accurately guides navigation from the start observation to the goal. Focus on essential actions and prominent, easily identifiable landmarks.

\textbf{Previous Instruction}: Walk and stop right before washing area.

\textbf{Input obs:}

Previous obs:
~\includegraphics[height=1.7cm]{ 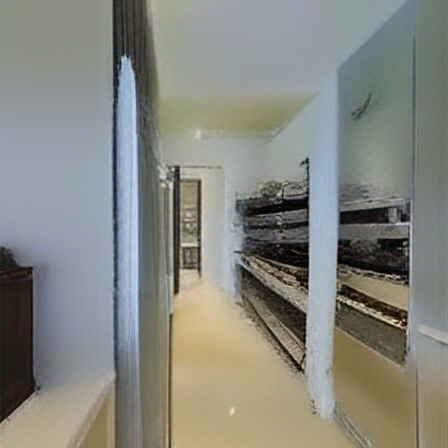}~\includegraphics[height=1.7cm]{ img/step_4_obs.png}


Current obs: 
~~~\includegraphics[height=1.7cm]{ img/step_5_obs.png}

Goal obs:
~~~~\,~~~\includegraphics[height=1.7cm]{ img/goal_obs.png}

\textbf{\tcbox[colback=gray!20, colframe=gray!20, boxrule=0.1pt,top=0mm, bottom=0mm]{~~~~~~~~~~~~~~~~~~~~~~~~~~~~~~~~~~~~Response~~~~~~~~~~~~~~~~~~~~~~~~~~~~~~~~~}}
\textbf{Predicted Instruction:}

Turn right and continue down the hall until you get to a refrigerator.
\end{fullpromptbox}
\caption{\small Prompt design examples on Interleaved multimodal reasoning during inference for Instruction Generation with Visual Cues. (Context size $m$ = 3)}
\label{fig:app_iter_ig}
\vspace{-0.2cm}
\end{figure}

\noindent\textbf{Interleaved Multimodal Reasoning.} As shown in Fig.~\ref{fig:app_iter_nv}, the model conditions each visual prediction on current/past observations, enhancing adaptability in dynamic or ambiguous scenes. For instruction refinement (Fig.~\ref{fig:app_iter_ig}), it evaluates the previously generated instruction against the updated visual context, revising it accordingly (e.g., “Turn right and continue down the hall until you reach a refrigerator”), thereby maintaining alignment with evolving scene cues.

\noindent\textbf{Comparison.} One-Pass reasoning prioritizes efficiency and simplicity, whereas Interleaved reasoning offers greater flexibility and robustness. The accompanying figures and outputs highlight how tailored prompt designs can elicit complementary strengths across multimodal navigation tasks.

\section{Experiments Details}
\label{appx-sec:exp-details}

\subsection{Evaluation Metrics}
We evaluate overall system performance using two complementary categories of metrics:

\noindent \textbf{(1) Instruction Quality:} Linguistic fidelity is comprehensively assessed for both goal-conditioned and visually grounded instruction generation using widely adopted text-generation metrics: BLEU-4~\cite{papineni2002BLEU}, CIDEr~\cite{vedantam2015cider}, METEOR~\cite{banerjee2005meteor}, and ROUGE-L~\cite{lin2004rouge}. Each generated instruction is compared against the full set of human-authored reference texts to ensure thorough and comprehensive coverage.

\noindent \textbf{(2) Visualization Quality:} For the navigation visualization subtask, visual predictions are evaluated with a combination of standard structural and perceptual measures, namely SSIM~\cite{wang2004ssim}, PSNR~\cite{hore2010image}, LPIPS~\cite{zhang2018unreasonable}, and DreamSim~\cite{fu2023dreamsim}. The latter two are deep perceptual metrics specifically designed to more closely approximate human judgments.

\noindent \textbf{LPIPS:} The Learned Perceptual Image Patch Similarity~\cite{zhang2018unreasonable} quantifies perceptual resemblance by computing weighted distances between deep feature activations extracted from pretrained vision backbones (e.g., AlexNet, VGG). By operating in a learned feature space, LPIPS better captures perceptually relevant differences than conventional low-level pixel-level measures.

\noindent \textbf{DreamSim:} DreamSim extends perceptual evaluation to the multimodal domain by measuring semantic alignment between generated images and a target text description. Given images \( \{I_i\}_{i=1}^{N} \) and a prompt \( T \), it is defined as:
\begin{equation}
\begin{aligned}
  \operatorname{DreamSim}&(I_{1:N}, T) =\\
  &\frac{1}{N}\sum_{i=1}^{N} \frac{\langle f_{\text{img}}(I_i),\, f_{\text{text}}(T) \rangle}{\|f_{\text{img}}(I_i)\|\cdot\|f_{\text{text}}(T)\|}\,.  
\end{aligned}
\end{equation}
Unlike the standard CLIP score, DreamSim leverages fused or fine-tuned visual–textual features (e.g., CLIP, OpenCLIP, DINO) trained on synthetic human similarity judgments, thereby further enhancing sensitivity to nuanced perceptual and semantic correspondences.

By combining LPIPS and DreamSim, our evaluation jointly accounts for low-level visual fidelity and high-level semantic coherence, offering a balanced and human-aligned assessment across both structural and semantic dimensions.

\subsection{Implementation Details for SOTA Methods}

In this section, we provide further implementation details on the SOTA navigation instruction generation methods we compare in Table 1 of main text.

\noindent \textbf{Speaker-Follower:} The original work \cite{fried2018speakerfollowermodelsvisionandlanguagenavigation} uses a speaker-follower architecture for vision-and-language navigation, where a follower maps instructions to actions and a speaker generates instructions from routes, enabling data augmentation and pragmatic inference with panoramic action space. We modified it to process sequential egocentric RGB observations, using ResNet-152 to encode input observations ($\mathcal{O}_{full} = \{o_1, \dots, o_k, o_g\}$) and an LSTM decoder with an attention mechanism for instruction generation. We remove the panoramic action space and adapt the model to work with first-person visual observations only.

\noindent \textbf{LANA:} Adapted from \cite{wang2023lanalanguagecapablenavigatorinstruction} by extracting its instruction generation module. The original work takes navigation routes (panoramic observations and actions) as input and generates natural language instructions as output, using a unified architecture with shared route/language encoders and cross-attention based decoders for bidirectional translation, jointly trained on both instruction following and generation tasks. We replace the panoramic encoder with ViT-based image encoding. Processes input observations ($\mathcal{O}_{full}$) through cross-attention, removing dependencies on privileged inputs (trajectory coordinates, maps, action labels).

\noindent \textbf{GPT-4o Direct (Zero-shot):} Processes input observations ($\mathcal{O}_{full}$) through direct prompting. The model receives explicit instructions that images 1-k represent continuous observations from the starting point along the path, while goal image shows the goal destination. We enforce strict output constraints: (1) no reference to image numbers in the instruction, as the end user will not have access to these images; (2) pure text output without any markdown formatting, bullet points, or special symbols; (3) single continuous paragraph format; and (4) concise instructions  for single-scene navigation. We require concise output because other models and baselines are trained on scene-segmented tasks and naturally produce shorter predictions, while GPT-4o tends to generate longer, more detailed instructions due to the task complexity, which can dilute its true capabilities in certain evaluation metrics. Images are encoded as base64 and resized to a maximum of 512×512 pixels to optimize API usage. The model generates instructions using temperature=0.7 and top\_p=0.95 for balanced creativity and coherence.

\noindent \textbf{GPT-4o CoT (Zero-shot):} Extends the direct approach with structured chain-of-thought reasoning. The model follows a five-step analysis process: (1) describe the starting position and environment, (2) identify key landmarks and direction changes, (3) describe the path progression, (4) identify the destination, and (5) generate the final navigation instruction. The same output constraints apply as the direct method, with the additional requirement that the final instruction must be prefixed with "FINAL INSTRUCTION:" for automatic extraction. This allows the model to perform detailed visual analysis while ensuring the final output remains concise. The complete reasoning process is preserved for analysis, while only the extracted final instruction is used for evaluation.

\noindent \textbf{C-Instructor:} Following \cite{Kong_2024}, which takes navigation trajectories with panoramic observations (36 views per step) and actions as input, using Chain-of-Thought with Landmarks (CoTL) to extract critical landmarks before instruction generation and Spatial Topology Modeling Task (STMT) for enhanced spatial understanding. We adapt their method for egocentric observations, using Llama-2-7B with CLIP-ViT-L-14 (36 patches/image) to process input observations ($\mathcal{O}_{full}$). The CoTL mechanism is modified for egocentric views instead of panoramic observations.

\begin{figure*}[!t]
\setlength{\abovecaptionskip}{2pt}
\setlength{\belowcaptionskip}{0pt}
\centering
    \includegraphics[width=\textwidth]{ 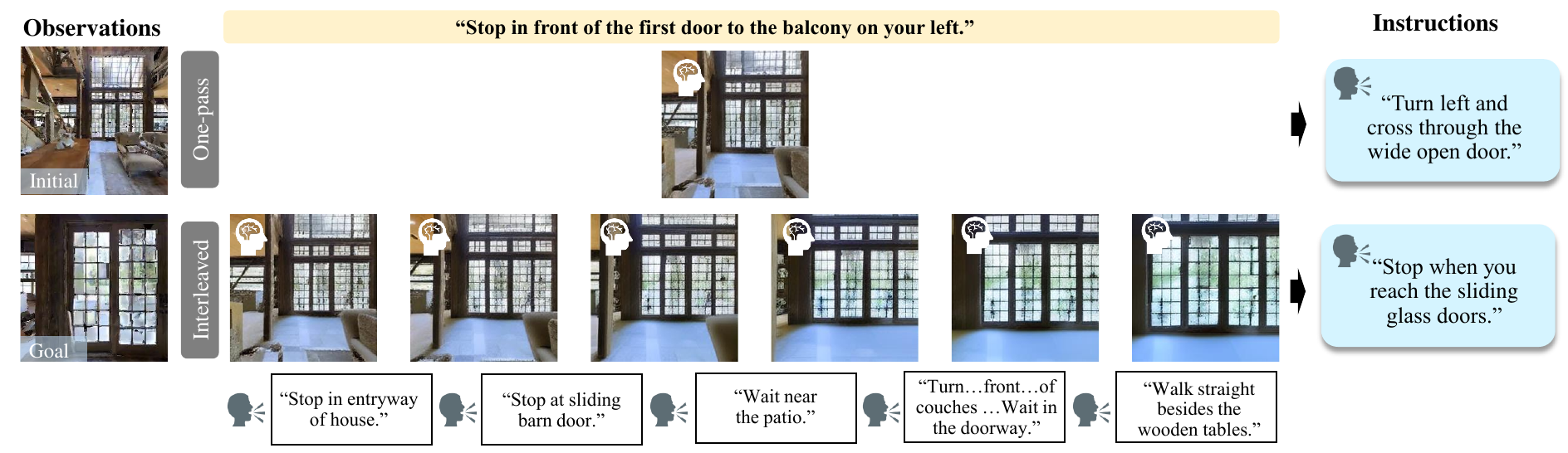}
\caption{\small \textbf{More qualitative results} on R2R-Goal unseen split (with ground truth) of our One-pass and Interleaved Reasoning.}
\label{fig:app_unseen}
\vspace{-0.2cm}
\end{figure*}

\noindent \textbf{Anole-7B Direct (Zero-shot):} We employ Anole-7B in a zero-shot setting with sparse observation sampling due to token constraints (4096 tokens total, 1024 tokens per image): $\mathcal{O} = \{o_1, o_k, o_g\}$. As Anole is primarily designed as an image generation model, it requires exceptionally detailed task specifications and comprehensive natural language descriptions to pass the model's regulation. We provide extensive task context explaining: the navigation instruction generation objective, the specific role of each observation (initial position at frame 1, final observation before turning at frame $k$, and goal destination), and explicit generation requirements for producing clear, actionable instructions. This detailed prompting approach compensates for Anole's architectural expectations without introducing any additional task-specific information beyond what other models receive—the enhancement lies solely in the completeness and clarity of the natural language task description.

\noindent \textbf{Anole-7B CoT (Fine-tuned):} We fine-tune Anole-7B using the CoT reasoning approach, wherein the model learns to generate navigation instructions through structured reasoning steps. These include (1) analyzing and describing the visual content of key observations, (2) identifying spatial relationships and environmental changes between the initial and final frames, (3) reasoning about the navigation trajectory from start to goal, and (4) synthesizing these elements into coherent instructions. Unlike the zero-shot setting, the fine-tuned model no longer relies on explicit task specifications or detailed natural language prompts. Through training with ground truth divided instructions, Anole-7B effectively internalizes both the objective and the reasoning patterns required for high-quality instruction generation.

\subsection{Implementation Details for SOTA Navigation Visualization Methods}

\noindent \textbf{GPT-4o + DALL·E:} We implement the two-stage approach for navigation visualization. Given three observations (previous $o_{t-1}$, current $o_t$, and goal $o_g$), GPT-4o analyzes the visual context and generates a text prompt describing the expected next observation $\hat{o}_{t+1}$. This prompt is then passed to DALL·E 2 for image synthesis. The system processes 3 input images and generates 1 output image per prediction.

\subsection{Prompt Design and Examples}
In this section, we provide prompt examples of our implementation on LLM-related SOTA methods.

\noindent \textbf{GPT-4o Direct Prompt:}
\begin{lstlisting}[language={}, breaklines=true, basicstyle=\small\ttfamily]
Task: Navigation Instruction Generation

You are given 7 images from a navigation trajectory:
- Images 1-6: Continuous observations from the starting point along the path
- Image 7: The goal/destination point

Generate a clear navigation instruction that guides someone from the starting point to the goal.

IMPORTANT REQUIREMENTS:
1. Do not reference image numbers (e.g., 'Start at the point shown in Image 1') in your instruction. The person receiving your instruction will not have access to these images. Describe locations and landmarks directly instead.
2. Output ONLY plain text. Do not use markdown formatting, bullet points, numbered lists, bold text (**text**), headers (#), or any other formatting symbols.
3. Write your instruction as a single continuous paragraph.
4. Since the navigation target is within a single scene, please make your instruction more concise.
\end{lstlisting}

\noindent \textbf{GPT-4o Chain-of-Thought Prompt:}
\begin{lstlisting}[language={}, breaklines=true, basicstyle=\small\ttfamily]
Task: Navigation Instruction Generation 

You are given 7 images from a navigation trajectory:
- Images 1-6: Continuous observations from the starting point along the path
- Image 7: The goal/destination point

Please analyze step by step:
1. Describe the starting position and environment
2. Identify key landmarks and direction changes
3. Describe the path progression
4. Identify the destination
5. Generate a clear navigation instruction

IMPORTANT REQUIREMENTS:
1. In your final navigation instruction, do not reference image numbers (e.g., 'Start at the point shown in Image 1'). The person receiving your instruction will not have access to these images. Describe locations and landmarks directly instead.
2. Use ONLY plain text throughout your response. Do not use markdown formatting, bullet points, numbered lists, bold text (**text**), headers (#), or any other formatting symbols.
3. Write your analysis and final instruction as continuous paragraphs.
4. For the final navigation instruction (step 5), since the navigation target is within a single scene, please make it more concise.
5. You MUST prefix your final navigation instruction with 'FINAL INSTRUCTION:' on a new line.
\end{lstlisting}

\noindent \textbf{C-Instructor Prompt:}
\begin{lstlisting}[language={}, breaklines=true, basicstyle=\small\ttfamily]
Based on these 7 navigation images showing a path (6 consecutive observations + 1 destination), analyze the scene step by step:

1. First, identify key objects and landmarks in each image:
[IMAGE_TOKEN] [IMAGE_TOKEN] [IMAGE_TOKEN] [IMAGE_TOKEN] [IMAGE_TOKEN] [IMAGE_TOKEN] [IMAGE_TOKEN]

2. Next, perceive the spatial relationships and transitions between consecutive frames:
- How does the viewpoint change from one frame to the next?
- What directional movements (forward, turn left/right) are implied?
- Which landmarks remain visible across multiple frames?

3. Finally, generate a clear and complete navigation instruction that guides someone from the starting point (image 1) to the destination (image 7):
\end{lstlisting}

\noindent \textbf{Anole CoT Finetuned Prompt:}
\begin{lstlisting}[language={}, breaklines=true, basicstyle=\small\ttfamily]
Task: Generate navigation instruction based on key observations.
You are given three key observations from a navigation path:
1. Initial observation at starting point: <image>
   Description: [GENERATED_DESCRIPTION_1]
2. Final observation at starting point (before turning): <image>
   Description: [GENERATED_DESCRIPTION_2]
3. Goal observation at destination: <image>
   Description: [GENERATED_DESCRIPTION_3]

Based on these observations, analyze step-by-step:
1. First, identify the key landmarks and spatial layout at the starting point.
2. Next, determine the navigation direction and movement pattern by comparing the initial and final observations at the starting point.
3. Then, analyze the goal observation to understand the target location and its distinguishing features.
4. Finally, synthesize a clear and complete navigation instruction that guides from the starting point to the destination.
Instruction:
\end{lstlisting}

\noindent \textbf{GPT-4o + DALL·E Navigation Visualization Prompt:}
\begin{lstlisting}[language={}, breaklines=true, basicstyle=\small\ttfamily]
Task: Navigation Single Step Visualization
Description: Given three observations from the previous first-person observation, the current first-person observation, and the goal observation, respectively, predict the next first-person observation the agent would see if it continues toward the goal.
Input observations are Previous observation, Current observation, and Goal observation
\end{lstlisting}

\subsection{Visual Results}
We present further qualitative results to supplement the illustrative examples provided in the main manuscript. As shown in Fig.~\ref{fig:app_unseen} and Fig.~\ref{fig:app_social}, we include complete sequences of observations paired with their corresponding instructions, visualized for both the unseen subset and real-world environments. Compared with ground-truth annotations, the generated instructions reliably capture the majority of salient landmarks and key objects, while producing correct navigational actions. This holds consistently across challenging settings, including unseen scenes and cluttered real-world trajectories, thereby demonstrating the robustness and generalizability of our reasoning strategies.
\begin{table}[!t]
\renewcommand\arraystretch{1.05}
\setlength{\abovecaptionskip}{2pt}
\setlength{\belowcaptionskip}{2pt}
\setlength{\tabcolsep}{1.5mm}
\small
\centering
\resizebox{\linewidth}{!}{
\begin{tabular}{cccccc}
\toprule
\textbf{Context}&\textbf{Token Len.}&\textbf{SSIM $\uparrow$}&\textbf{PSNR $\uparrow$} &\textbf{LPIPS $\downarrow$} &\textbf{DreamSim $\downarrow$}\\
\midrule
1 & 784 & 0.66 & 19.20 & 0.29 & 0.15 \\
\rowcolor{blue!10}2 & 784 & \textbf{0.69} & \textbf{20.02} & \textbf{0.27} & \textbf{0.13} \\
4 & 400 & 0.55 & 16.30 & 0.35 & 0.19 \\
\bottomrule
\end{tabular}}
\caption{\small \textbf{Trade-off} between Context Size and Image Token Length on Navigation Visualization (val unseen). Token Length denotes visual token number per frame.}
\label{tab:anole7b_context_size}
\vspace{-0.1cm}
\end{table}
\begin{table}[!t]
\renewcommand\arraystretch{1.05}
\setlength{\abovecaptionskip}{2pt}
\setlength{\belowcaptionskip}{2pt}
\small
\centering
\setlength{\tabcolsep}{1mm}
\resizebox{\linewidth}{!}{
\begin{tabular}{cccccc}
\toprule
\textbf{Context}&\textbf{Token Len.}&\textbf{BLEU-4 $\uparrow$}&\textbf{CIDEr $\uparrow$}&\textbf{METEOR $\uparrow$}&\textbf{ROUGE-L $\uparrow$}\\
\midrule
2 & 784 & 0.26 & 0.15 & 0.14 & 0.16 \\
\rowcolor{blue!10}3 & 784 & \textbf{0.30} & \textbf{0.18} & \textbf{0.17} & \textbf{0.20} \\
5 & 400 & 0.22 & 0.13 & 0.11 & 0.13 \\
\bottomrule
\end{tabular}}
\caption{\small \textbf{Trade-off} between Context Size and Image Token Length on instruction generation with visual cues.(val unseen)}
\label{tab:anole7b_instruction_context}
\vspace{-0.1cm}
\end{table}

\subsection{Detailed Ablation Results}
We analyze the influence of context size and image token length in Figure~\ref{fig:content}. The detailed values on more metrics are provided in Tables~\ref{tab:anole7b_context_size} and \ref{tab:anole7b_instruction_context}.

{
    \small
    \bibliographystyle{ieeenat_fullname}
    \bibliography{main}
}

\end{document}

%% file: preamble.tex
\usepackage{amsmath, amssymb, bm, bbm, mathrsfs}
\usepackage{siunitx}

\usepackage[utf8]{inputenc}
\usepackage[T1]{fontenc}
\usepackage{microtype}

\usepackage[dvipsnames]{xcolor}
\usepackage{soul}
\setul{0.5ex}{0.3ex}

\usepackage{booktabs, tabularx, threeparttable, multirow, array, colortbl, makecell}
\usepackage{tabularray}
\UseTblrLibrary{booktabs}

\usepackage{enumitem}
\setlist[enumerate]{noitemsep, topsep=0pt}
\setlist[itemize]{noitemsep, topsep=0pt}

\usepackage{caption}
\usepackage{subcaption}

\usepackage{algorithm}
\usepackage{algorithmic}

\usepackage{url}
\usepackage{pifont}
\usepackage{setspace}
\usepackage{wrapfig}
\usepackage{tikz}
\usetikzlibrary{positioning}
\usepackage[most]{tcolorbox}
\usepackage{wasysym}

\makeatletter
\newcommand{\thickhline}{%
    \noalign {\ifnum 0=`}\fi \hrule height 1pt
    \futurelet \reserved@a \@xhline
}
\makeatother